\pdfoutput=1

\documentclass[11pt]{article}

\usepackage[]{acl}

\usepackage{times}
\usepackage{latexsym}

\usepackage[T1]{fontenc}

\usepackage[utf8]{inputenc}

\usepackage{microtype}
\usepackage{graphicx}
\usepackage{paralist}
\usepackage{multirow}
\usepackage[export]{adjustbox}
\usepackage{booktabs}
\usepackage{nicematrix}
\usepackage{xcolor}
\usepackage{colortbl}
\usepackage[outline]{contour}
\usepackage{enumitem}
\usepackage{siunitx}
\usepackage{subcaption}
\usepackage{xpinyin}
\usepackage{CJKutf8}
\usepackage{threeparttable}
\usepackage{graphicx}
\usepackage{caption}

\newcommand{\py}[1]{\begin{CJK}{UTF8}{gbsn}\xpinyin*{#1}\end{CJK}}

\newcommand{\ku}{$^1$}
\newcommand{\wa}{$^2$}
\newcommand{\com}{$^3$}
\newcommand{\cuhksz}{$^4$}
\newcommand{\sribd}{$^5$}
\newcommand{\uva}{$^6$}

\title{FoodieQA: A Multimodal Dataset for Fine-Grained \\ Understanding of Chinese Food Culture}

\author{Wenyan Li,\ku~Xinyu Zhang,\wa~Jiaang Li,\ku~Qiwei Peng,\ku ~Raphael Tang,\wa$^,$\com~Li Zhou,\cuhksz$^,$\sribd ~\textbf{Weijia Zhang},\uva\\
\textbf{Guimin Hu},\ku 
~\textbf{Yifei Yuan},\ku
~\textbf{Anders Søgaard},\ku
~\textbf{Daniel Hershcovich},\ku
~\textbf{Desmond Elliott}\ku \\
{\ku}University of Copenhagen {\wa}University of Waterloo {\com}Comcast AI Technologies\\
{\cuhksz}The Chinese University of Hong Kong, Shenzhen
{\sribd}Shenzhen Research Institute of Big Data \\
{\uva}University of Amsterdam\\
\texttt{weli@di.ku.dk}
}

\begin{document}
\maketitle

\begin{abstract}
Food is a rich and varied dimension of cultural heritage, crucial to both individuals and social groups. To bridge the gap in the literature on the often-overlooked regional diversity in this domain, we introduce FoodieQA, a manually curated, fine-grained image-text dataset capturing the intricate features of food cultures across various regions in China. We evaluate vision--language Models (VLMs) and large language models (LLMs) on newly collected, unseen food images and corresponding questions. FoodieQA comprises three multiple-choice question-answering tasks where models need to answer questions based on multiple images, a single image, and text-only descriptions, respectively. While LLMs excel at text-based question answering, surpassing human accuracy, the open-weights VLMs still fall short by 41\% on multi-image and 21\% on single-image VQA tasks, although closed-weights models perform closer to human levels (within 10\%). Our findings highlight that understanding food and its cultural implications remains a challenging and under-explored direction.

\end{abstract}

\section{Introduction}
\label{sec:intro}
\begin{figure}[t]
        \includegraphics[width=\columnwidth]{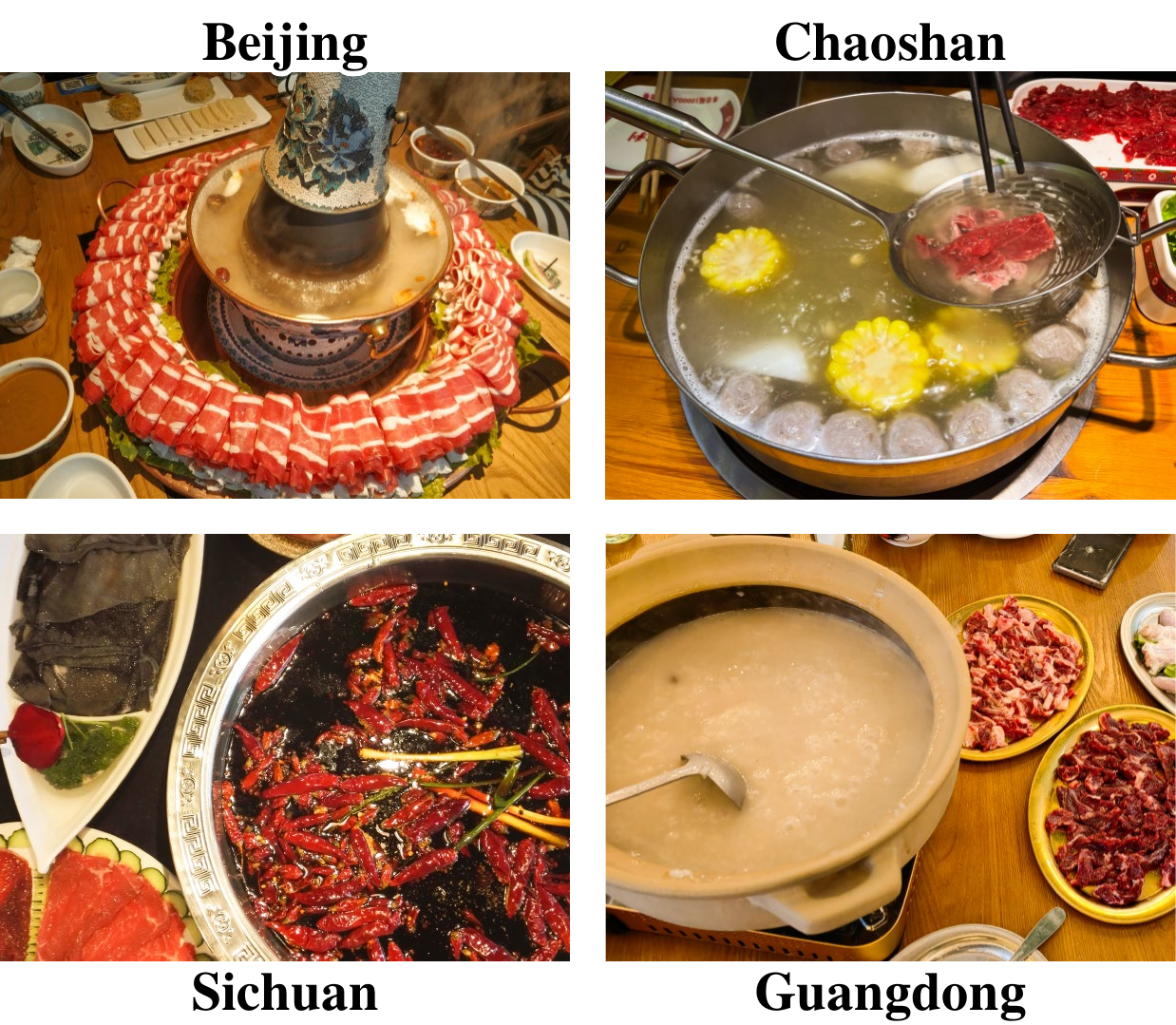}
        \caption{An example of regional food differences in referring to \textit{hotpot} in China. The depicted soups and dishware visually reflect the ingredients, flavors, and traditions of these regions: Beijing in the north, Sichuan in the southwest, and Guangdong in the south coast.\label{fig:hotpot}}
    \end{figure}

One of the most popular dishes %
in China is {\em hotpot}, which comes in many varieties, as shown in Figure~\ref{fig:hotpot}:\ Beijing is renowned for its mutton hotpot served with a traditional copper pot (\begin{CJK}{UTF8}{gbsn}\xpinyin*{铜锅涮羊肉}\end{CJK}).  Guangdong province is home to a famous porridge-based hotpot (\begin{CJK}{UTF8}{gbsn}\xpinyin*{粥底火锅}\end{CJK}), while its coastal region of Chaoshan is known for beef hotpot (\begin{CJK}{UTF8}{gbsn}\xpinyin*{潮汕牛肉火锅}\end{CJK}). %
The hotpot varieties from Sichuan and Chongqing are celebrated for their flavorful broths, with chili peppers and Sichuan peppercorns that create a unique numbing-spicy sensation. 
The variation among regional cultures within a country %
highlights the challenges that language models face in understanding cultural knowledge and context-specific information in the food domain. 

Existing datasets and models that focus on food and culinary practices primarily concentrate on tasks such as food recognition, recipe generation, food knowledge probing or recipe-related question answering~\cite{chen2017chinesefoodnet, cao_cultural_2024, zhou2024does, yagcioglu_recipeqa_2018}. However, they often take a coarse view, conflating country, culture and language. Important regional cultural differences remain under-studied \cite{palta_fork_2023}.

We introduce \textbf{FoodieQA}, a manually curated set of multimodal test questions designed to probe fine-grained cultural awareness with a focus on the food domain. Our dataset targets two under-explored directions:\ regional cultural diversity within a country and challenging fine-grained vision-language understanding in the culinary domain.

    \begin{figure*}[t]
    \centering
        \hspace{3.7mm}\includegraphics[width=0.9\textwidth, clip]{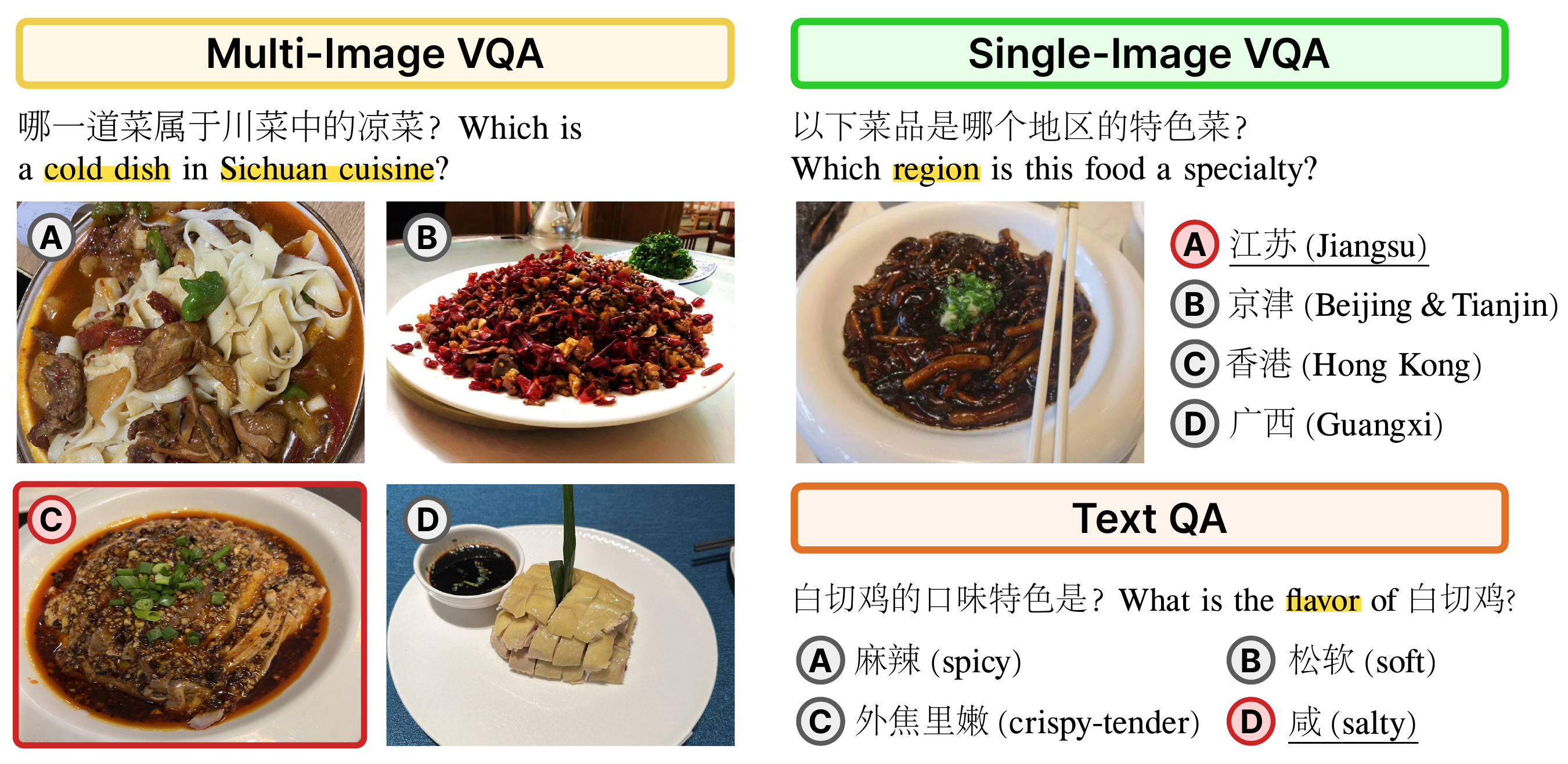}
        \caption{The tasks in FoodieQA evaluate food culture understanding from three perspectives. \textit{Multi-image VQA} requires the ability to compare multiple images, similar to how humans browse a restaurant menu. \textit{Single-image VQA} assesses whether models can use visual information to better understand food culture. 
        \textit{Text-based} questions probe model performance without multimodal data.\footnotemark Fine-grained attributes that the questions focus on are highlighted. \label{fig:ex}}
    \end{figure*}
    
To build a regionally diverse dataset, we gather dishes and images selected by native Chinese speakers from various regions, covering 14 distinct cuisine types across China. To ensure the images used for benchmarking are fresh and have no chance of leaking into the pretraining data of VLMs, we collect images uploaded by local people, which are not publicly available online. We then define multiple attributes associated with the dishes and have native Chinese annotators create multiple-choice questions based on their expertise. Our dataset includes both vision-based question answering and text-based question answering tasks, as illustrated in Figure~\ref{fig:ex}.

\footnotetext{We only evaluate TextQA in Chinese to prevent bias introduced through translating dish names. The English translation is only for illustration purpose.}

We benchmark a series of state-of-the-art models, including seven LLMs and eight VLMs, on the Foodie dataset using zero-shot evaluation. By comparing their performance to human accuracy, we highlight the gap between open-weights and closed-weights models and demonstrate their limitations in understanding Chinese regional food culture. Additionally, we compare the performance of bilingual models trained on both Chinese and English datasets to English-focused models, revealing biases in their understanding of region-specific food culture and the language of the questions. Finally, our analysis shows that visual information improves the performance of VLMs compared to text-only inputs, although some models struggle with identifying dishes from images.

\section{Related Work}
\label{sec:related}

\paragraph{Multilingual Multimodal Datasets} Multimodal systems are typically evaluated on English due to the widespread availability of English-language datasets. However, there are some examples of research on training and evaluating models beyond English for image captioning~\cite{elliott-etal-2016-multi30k}, image--sentence retrieval~\cite{10.1145/3404835.3463257}, visual reasoning~\cite{liu-etal-2021-visually}, and question-answering~\cite{pfeiffer-etal-2022-xgqa}. This paper focuses on Chinese visual question answering, with fine-grained attributes in the food domain.

\paragraph{Food Datasets}
In recent years, most food datasets have been designed for food image classification \citep{chen2017chinesefoodnet}, food captioning \citep{ma_food-500_2023}, and recipe-focused generation and question answering \citep{yagcioglu_recipeqa_2018,yummly66, liu2022counterfactual}. For culture knowledge probing in the food domain, some of the recent datasets span multiple countries and include broad cultural or regional metadata \citep{yummly66,ma_food-500_2023, romero2024cvqa}. However, they often use country as a proxy for culture, such as the country of origin for the food. For example, \citet{palta_fork_2023} introduced a test set to probe culinary cultural biases by considering US and non-US traditions, \citet{zhou2024does} construct a multicultural, multilingual dataset focusing on culinary knowledge,
and \citet{cao_cultural_2024} focuses on recipe transfer between Chinese and English. Investigating cultural differences within a country remains an under-explored area \citep{palta_fork_2023}.

\paragraph{Fine-grained Vision-Language Understanding}
\citet{bugliarello_measuring_2023} quantified the fine-grained vision-language understanding capabilities in existing models, focusing on aspects within the general domain. Later works focus on the culture understanding in VLMs~\cite{Liu2023TowardsER,Cao2024ExploringVC}.  However, current fine-grained VL datasets \citep{zhang2021vsr, parcalabescu-etal-2022-valse, winoground_2022_CVPR, svo_prob} are often framed as binary classification tasks, which limits their difficulty. Concurrently with our work, \citet{romero2024cvqa} and \citet{nayak2024benchmarkingvisionlanguagemodels} have created culturally-diverse question-answering datasets across multiple countries. Our multi-choice vision question answering dataset that focuses on Chinese regional differences aims to advance the boundaries of fine-grained understanding in the context of food and culture.

\section{FoodieQA: Dataset Annotation}
\label{sec:dataset}
China, with its expansive territory and long history, has cultivated rich and diverse food culture and traditions. Focusing on regional food culture differences, our dataset collection contains five distinct phases. 1) selection of cuisine types inside China; 2) collection of private images; 3) individual dish annotation; 4) visual question formulation; 5) text question formulation.

\begin{figure}[t]
    \centering
    \includegraphics[width=0.9\columnwidth, trim=1mm 1mm 1mm 6mm, clip]{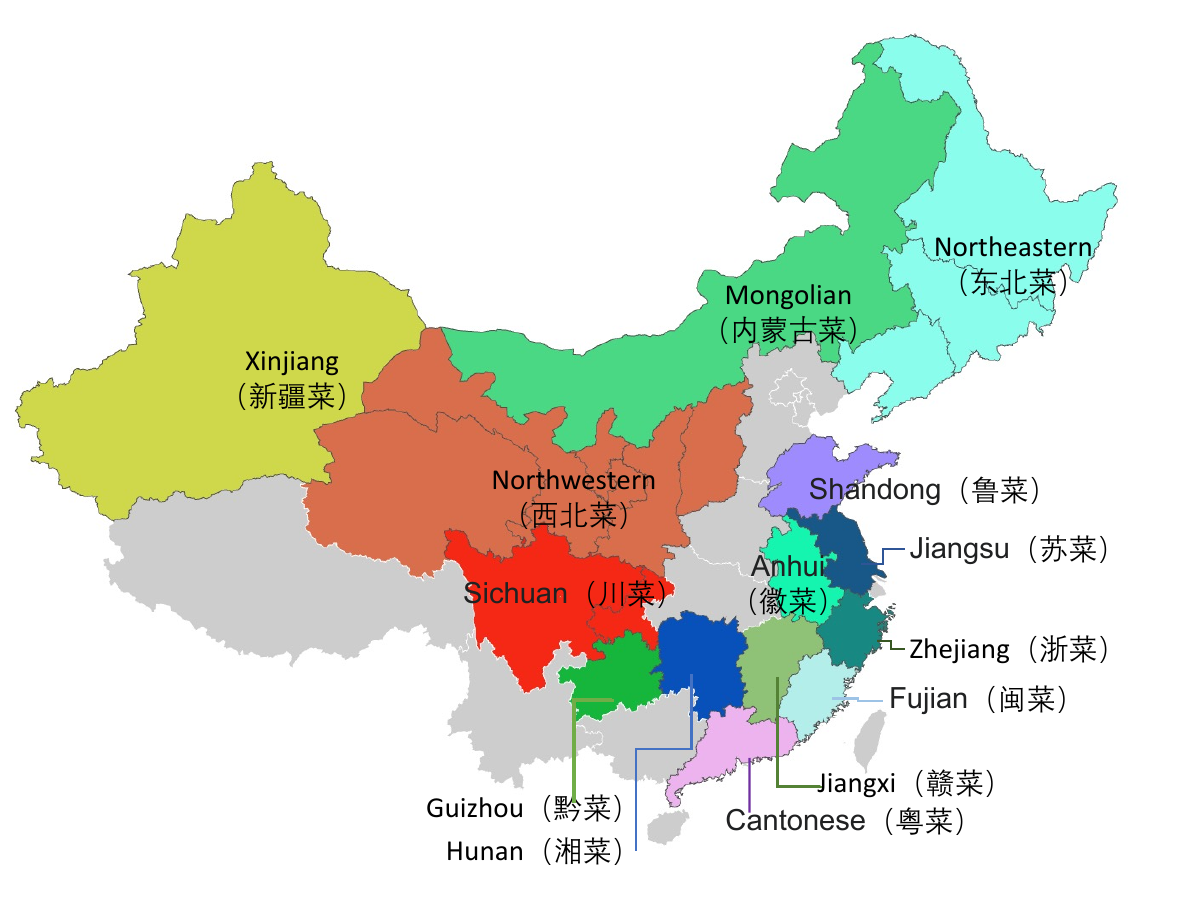}
    \caption{Geographical distribution of cuisine types.\footnotemark}
    \label{fig:cuisine-types}
\end{figure}
\footnotetext{We omit the Islands of the South China Sea in the figure for visualization simplicity.}

\subsection{Selection of Cuisine Types}
\label{subsec:cuisine-type} 
The well-recognized "eight major cuisines" in China are Sichuan (\py{川菜}), Guangdong (i.e., Cantonese, \py{粤菜}), Shandong (\py{鲁菜}), Jiangsu (\py{苏菜}), Zhejiang (\py{浙菜}), Fujian (\py{闽菜}), Hunan (\py{湘菜}), Anhui (\py{徽菜}) cuisines~\cite{eight-major}. This categorization is based on historical, cultural, and geographical factors that have influenced the development of distinct cooking styles and flavors in different regions of the country. For a better geographical coverage, we extend the eight cuisine types to additionally include Northwest (\py{西北菜}), Northeast (\py{东北菜}), Xinjiang (\py{新疆菜}), Jiangxi (\py{赣菜}) and, Mongolian cuisines (\py{内蒙古菜}) in this study. This results in 14 types (Figure~\ref{fig:cuisine-types}) in total, for which we collect dish images and annotations. 
    
\subsection{Collection of Images}
\label{subsec:image}
To ensure that the images are not used in the pretraining of existing models and contaminating evaluation, we designed and distributed a survey for Chinese locals to upload their own dish images (Figure~\ref{fig:uploads}).\footnote{The survey is distributed through WeChat and Douban.} We provide detailed guidelines for image uploading, specifying that: (1) the image should be clear, with a single dish as the focal point in the center; (2) participants should select the cuisine type of the dish from our list or specify it if it is not listed; (3) participants should provide the specific name of the dish, e.g., "mapo tofu (\py{麻婆豆腐})" instead of "tofu (\py{豆腐})"; (4) participants should indicate where the dish was served in their image, choosing from options such as cooked at home, restaurant, canteen, or delivery; (5) participants need to grant us permission to use the image for research purposes and confirm the image is not publicly available online, i.e., it has neither been downloaded from nor uploaded to the web or social media. In other words, the images we collected only existed on their phones or cameras. The uploaded images genuinely represent the locals' daily diet and culinary experiences, showcasing dishes that are currently popular.

We manually filter out 102 images that are blurry, have the dish off-center, or show a mismatch between the dish and the image. 
    \begin{figure}[t]
        \includegraphics[width=\columnwidth, trim=5mm 0mm 5mm 3mm, clip]{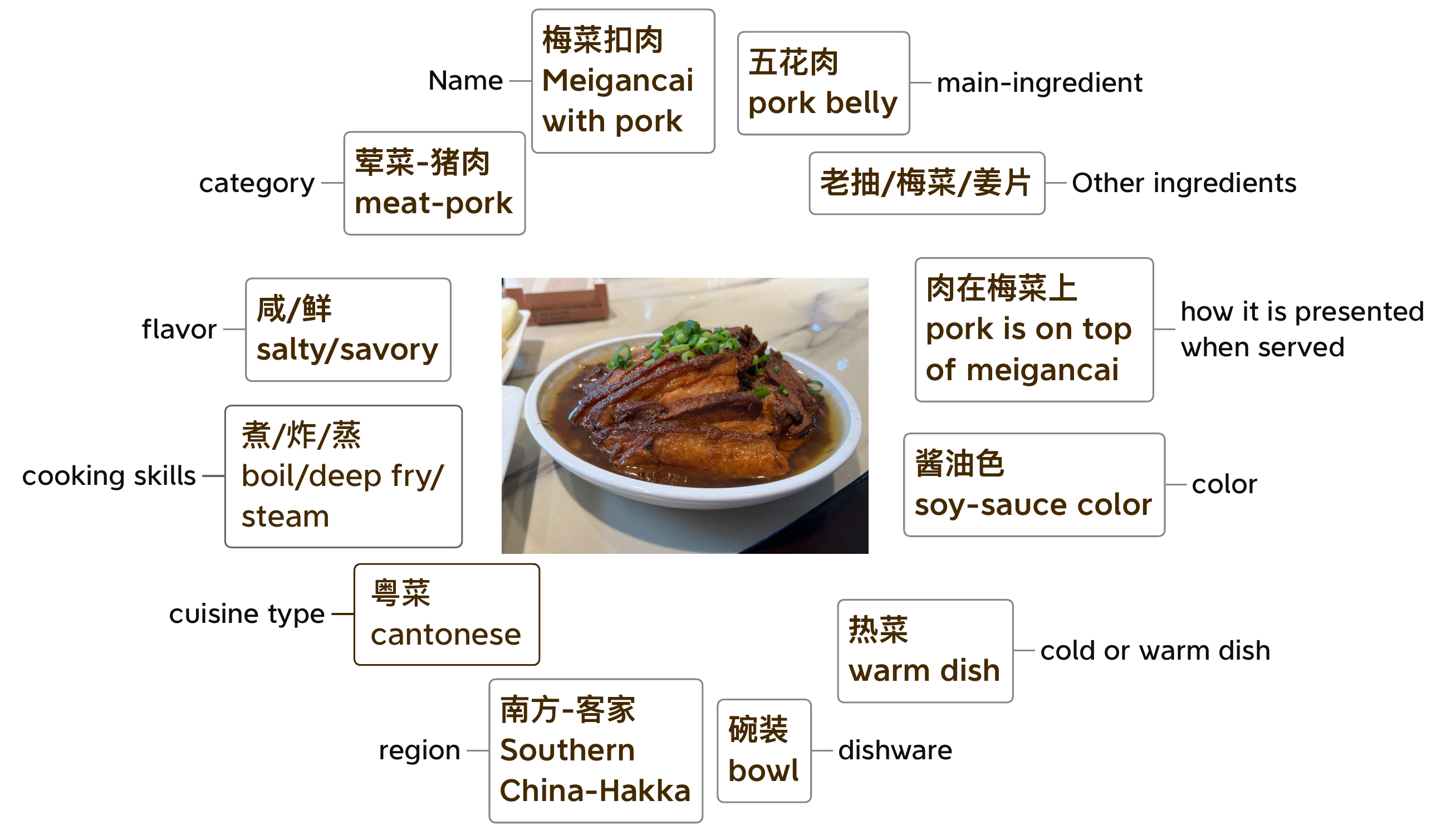}
        \caption{Meta-info annotation for local specialty.\label{fig:text-ann}}
    \end{figure}

\subsection{Local Specialty Annotation}
\label{subsec:dish-ann}
We also gather text annotations of representative local specialties for each cuisine type on our list. Annotators are asked to collect meta information for representative local dishes for each cuisine type, based on their life experience and knowledge obtained from the web. These meta-fields provide information beyond recipes, offering insights into how the food looks and tastes when people are eating it. An example is provided in Figure~\ref{fig:text-ann}. 

The 17 meta-info fields cover the appearance, taste, and culinary attributes of a dish. They include the food category, dish name, alternative names, main ingredient, characteristics of the main ingredient, three other key ingredients, dish flavor, presentation style, dish color, serving temperature (cold or warm), dishware used, region and province of origin, cuisine type, three primary cooking techniques, eating habits (if any), and reference links. 

The annotation is done by eight native Chinese speakers, including five PhD students and three postdoctoral researchers from various provinces in China.\footnote{The annotators are from Sichuan, Shaanxi, Guangdong, Jiangsu, Jiangxi, Shandong, and Chongqing.} During the annotation process, we ensure that all collected data is either annotated or verified by individuals familiar with the local context. Specifically, annotators are assigned as follows: 1) They are asked to annotate local specialties for the cuisine types from their hometowns, guaranteeing that the annotations are provided by locals. 2) If a local annotator can not be found for a specific cuisine type, annotators are requested to seek assistance from friends who are from the respective region to verify or correct the metadata obtained from the web. Annotations in the following sections are conducted by the same annotators, if not mentioned otherwise.

\subsection{Visual Question Answering Annotation}
\label{subsec:vqa}
One major consideration for vision-language understanding is that models can rely on  language priors, consequently neglecting visual information~\cite{vqav2, zhang2016yin}. This underscores the importance of formulating visual questions in such a way that they can only be answered by examining visual features, rather than relying on text priors. Based on the number of images used as inputs, we formulate both multi-image VQA questions and single-image VQA questions.

\subsubsection{Multi-image VQA}
\label{subsubsection:multi-image-vqa}
Multi-image VQA requires the ability to compare detailed visual features from multiple images, similar to how humans browse a restaurant menu.
\paragraph*{Question formulation}
We ask the annotators to write challenging questions that require: (1) looking at the dish images to answer, (2) thinking beyond merely recognizing the dish and questions that may require multi-hop reasoning, (3) asking diverse questions that belong to a diverse set of question types such as food type, flavor, color, expense, amount, and etc., (4) only one image is the correct answer to the question. The multi-image VQA questions are written by five native speakers from five different regions in China.

We organize the collected images into 28 groups based on cuisine types and food categories, as outlined in Section~\ref{subsec:image}. This allows annotators to write questions sequentially for related images extracted from the same group. Each annotator is asked to write two--three questions, given a four-image group. We note that in order to avoid the bias from language priors, dish names corresponding to the images are not presented. The user interface that we use for annotation is shown in Figure~\ref{fig:multi-vqa}.

\paragraph*{Question verification}
Once the questions and answers for the multi-image multiple-choice questions are collected, we verify the questions by asking the annotators (who did not create the questions) to answer them. If a question does not meet our defined criteria, annotators are instructed to flag it as a "bad question". Through this process, 87 questions were discarded. Additionally, when answering the questions, annotators are required to provide the rationale they use to reach the answer, as well as judge whether the question requires multi-hop reasoning. The user interface that we use for verification is shown in Figure~\ref{fig:multi-vqa-verify}. Each question is verified by two annotators, and we exclude the questions that do not have full agreement.

\subsubsection{Single-Image VQA}
\label{subsubsec:single-image-vqa}
Besides using images as multiple-choice answer options, we also ask diverse fine-grained questions about various aspects of a dish based on its meta-information (collected in Section~\ref{subsec:dish-ann}). We identify dishes that have both meta-information annotations and collected images, and then create questions based on the meta-information. As shown in the example in Figure~\ref{fig:ex}, the dish name is intentionally omitted from the questions to ensure they can only be answered by examining the visual features.

\paragraph*{Question formulation} 
We adopt a template-based approach, where a question about the same meta-field is asked multiple times, varying factors like the image of the dish, while the answer options are carefully selected from the wrong candidates in the meta-field to ensure that only one answer is correct. The single-image VQA questions are generated using a rule-based method, followed by thorough human verification that is similar to the multi-image VQA verification process. Please see details in the Appendix~\ref{sec:app-rules}.

\paragraph*{Question verification}
Similar to verification for the multi-image VQA questions, annotators are asked to answer the question given the text query and the corresponding image, and raise a "bad question" flag to filter out questions that does not satisfy the criteria. 88 questions were discarded as bad. Note that the name of the dish is not revealed in the text question so that the question needs to be answered based on visual information. Annotators are asked to write "I don't know" in the rationale and randomly guess an answer if they think the question is beyond their knowledge.

\subsection{Text Question Answering Annotation}
We formulate the text-based questions by combining human annotations and rule-based generation. Similar to the single-image VQA approach described in Section~\ref{subsubsec:single-image-vqa}, we generated questions and multiple-choice answer options based on the meta-information fields. However, instead of using the dish image, we included the dish name directly in the question. The questions are formulated using templates, where only the dish names and meta-fields are varied. A same human verification process to single-image question answering is included. 135 bad questions were discarded. Notice that annotators were asked to answer the questions based on their knowledge without using search engines, this makes the task challenging as it would be hard for one to answer questions about unfamiliar foods and regions without any other available information besides names of the food.

\label{subsec:qa}

\section{Dataset Statistics}
\label{sec:stats}

\subsection{Human Validation}
\label{subsec:human-validation}
In Table~\ref{tab:qstats}, we calculate human accuracy and inter-annotator agreement scores based on human-verified questions, excluding those identified as bad questions. For the single-image VQA and text QA questions, given the diverse cultural backgrounds of the human annotators, some questions can be challenging if the required food culture knowledge falls outside an annotator's cultural experience. For those questions, annotators are instructed to indicate "I don't know" and randomly guess an answer, as one might not be familiar with all of the specific dishes or the fourteen cuisine types. These questions are marked as out-of-domain. Considering the randomly selected answers for these out-of-domain questions allow us to obtain lower bound agreement and human accuracy scores.\footnote{Note that this is the only impact of the randomization. The ground truth label is annotated at an earlier stage of question formulation where the questions and choices are generated using the rule-based method.} We also report Cohen's Kappa ($\kappa$) and human accuracy separately for in-domain questions. 

The human validation process involves three postdoctoral researchers and five PhD students who are native Chinese speakers as introduced in Section~\ref{subsec:dish-ann}. Each question is verified and answered by two annotators who were not involved in the question formulation.  We retain the out-of-domain questions for calculating human accuracy and later in evaluating model performance, as the lower agreement scores are only due to differences in the annotators' cultural knowledge~\cite{plank-2022-problem}.

\begin{table}[!t]
    \centering
    \resizebox{\linewidth}{!}{
    \begin{tabular}{lccc}
          Task & Questions  & $\kappa$ & Accuracy \\
        \midrule
       Multi-image VQA & 403 & .834 & .916 \\
       \midrule
       Single-image VQA & 256 & .556 & .744 \\
       - In-domain & 168 & .674 & .818 \\
       \midrule
       Text QA & 705 & .470 & .562 \\
       - In-domain & 307 & .808 & .857 \\
        \bottomrule
    \end{tabular}
    }
    \caption{Statistics per task in FoodieQA. \label{tab:qstats}}
\end{table}

\begin{table}[t!]
    \centering
    \resizebox{\linewidth}{!}{
    \begin{tabular}{lccc}
           & Multi-image & Single-image & TextQA \\
        \toprule
       Avg. length & 12.9 & 17.0 & 14.9 \\
       Question types & 14 & 6 & 7 \\
       Multi-hop (\%) & 25.3 & 73.4 & 1.6\\
       Unique Images & 389 & 103 & - \\
        \bottomrule
    \end{tabular}}
    \caption{Question statistics. \label{tab:qstats2}}
\end{table}

\begin{figure}[t]
    \includegraphics[width=\columnwidth, trim=0mm 10mm 0mm 18mm, clip]{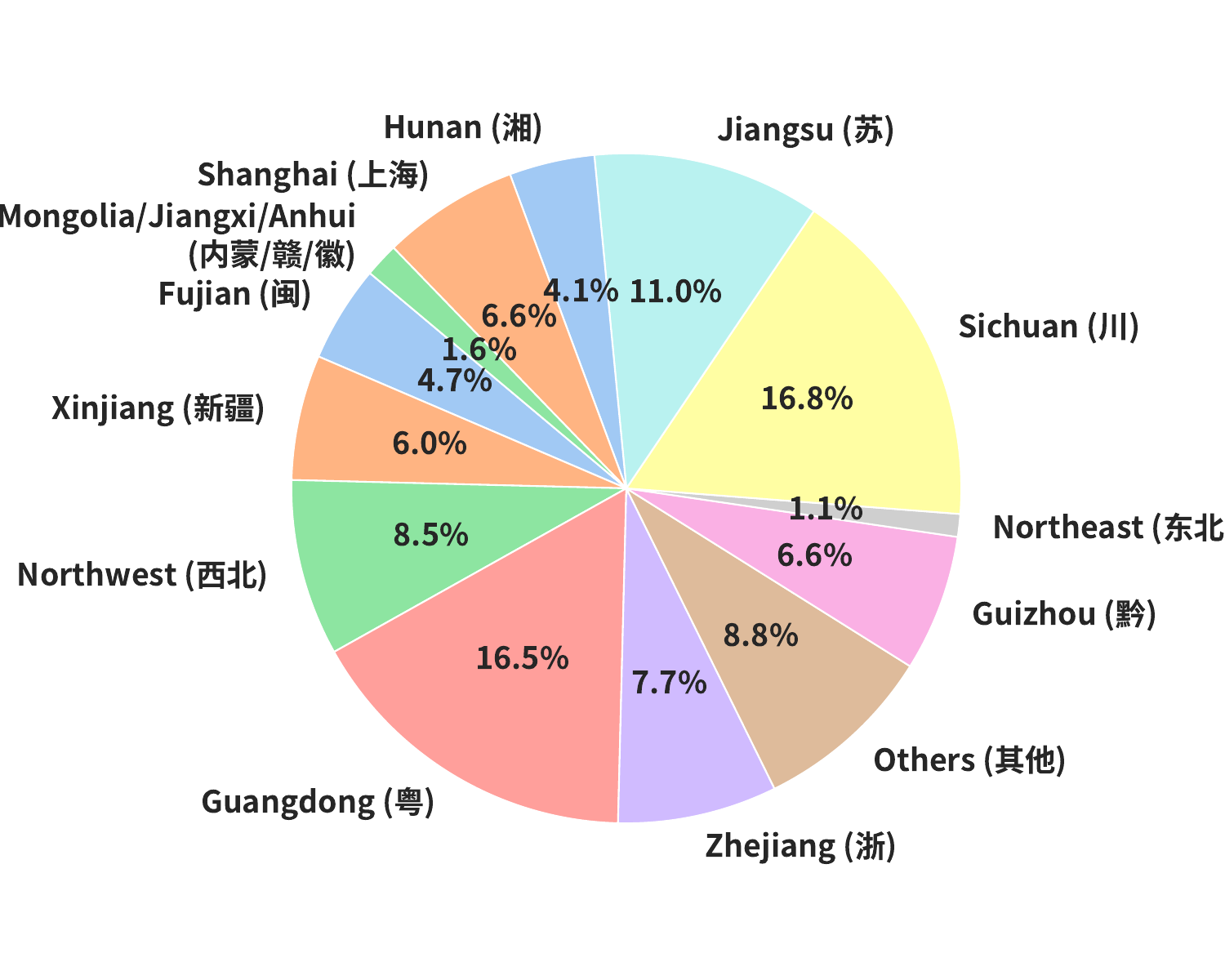}
    \caption{Region distribution of collected food images.\label{fig:img-distr}}
\end{figure}

\subsection{Image and Question Distribution}
\label{subsec:distr}

\paragraph{Image statistics}
\label{para:imgstats}

We collected 502 images but discarded 113 due to quality control issues. The final dataset of 389 images are distributed across regions in China as shown in Figure~\ref{fig:img-distr}. All 389 images are used for multi-image VQA; a subset of 103 images are used for single-image VQA.

\paragraph{Question statistics}
\label{para:qstats}
After human verification, we obtain 403 multi-image VQA questions, where each question needs to be answered with a set of four provided images. Single-image VQA tasks consists of 256 question in total, and text QA consists of 705 questions in total (Table~\ref{tab:qstats}). A considerable number of the VQA questions require multi-hop reasoning to predict the correct answer. We report the key statistics of the questions in Table~\ref{tab:qstats2}. Please see more details in Appendix~\ref{sec:app-stats}.

\section{Baselines:\ How Much of a Foodie are the LLMs/VLMs?}
We evaluate open-weight and API-based state-of-the-art LLMs and VLMs to probe their culture knowledge in the food domain. We evaluate the models in both Chinese and English for the VQA tasks. The questions are translated to English using the DeepL free API\footnote{\url{https://www.deepl.com/en/translator}} and validated by two PhD students who are Chinese native speakers and fluent in English. To avoid bias in translating dish names, we conduct the TextQA task solely in Chinese.

\subsection{Multi-Image VQA is Difficult}

\label{subsec:eval-mivqa}
We evaluate the multi-image VQA task using open-weight models that are capable of handling multiple image inputs, including Phi-3-vision-128k-instruct~\cite{abdin2024phi3}, Idefics2-8B~\cite{idefics2}, Mantis-8B-Idefics2~\cite{Jiang2024MANTISIM}, and English-Chinese bilingual Qwen-VL-12B~\cite{bai2023qwenvl}, and Yi-VL 6B and 34B models~\cite{ai2024yi}, as well as API-based models GPT-4V and GPT-4o~\cite{achiam2023gpt}. 

We experimented with four different prompts that utilized lists of images and texts or interleaved image-text inputs. Details can be found in Appendix~\ref{sec:app-prompts}. As shown in Figure~\ref{fig:mivqa-acc}, when compared to the human accuracy of 91.69\% in Chinese, the best-performing open-weight model, Idefics2-8B, achieves an accuracy of 50.87\%, which is still significantly lower than human performance. This indicates that current state-of-the-art models are still weak at distinguishing differences among food from visual input. This underscores that multi-image understanding, especially in contexts requiring cultural knowledge in the food domain, remains a challenging problem. When evaluating on the translated English questions, model performance decreases for all models except Phi-3-vision.

\begin{figure}[t]
    \centering
    \includegraphics[width=\columnwidth,trim=0mm 8mm 0mm 7mm, clip]{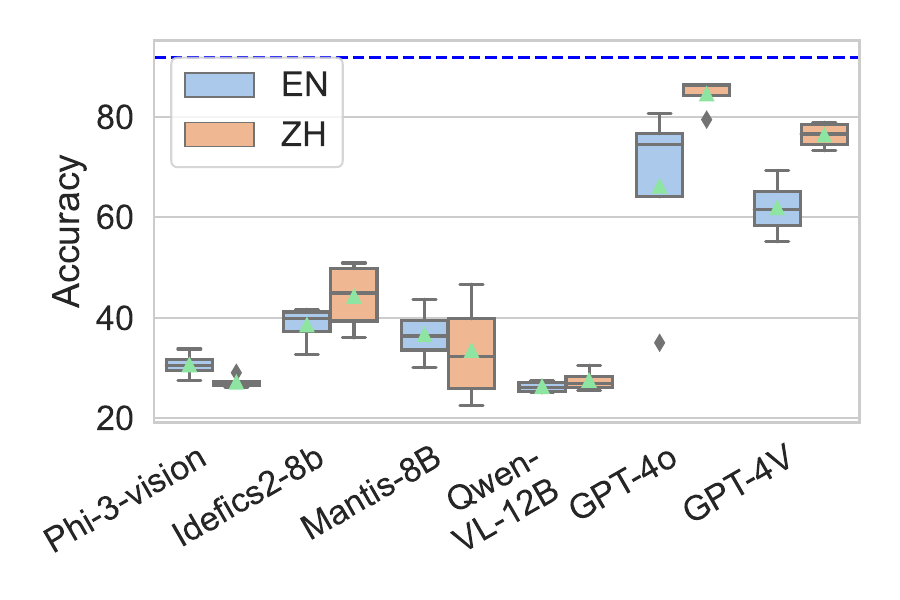}
    \caption{Accuracy of multi-image VQA tasks across four different prompts compared to a 91.96\% human accuracy in Chinese. Although Idefics2 and Mantis have higher accuracy than other models, they show greater variation across different prompts.}
    \label{fig:mivqa-acc}
\end{figure}

\subsection{Single-Image VQA Results}
Besides the four open sourced models that we used for multi-image VQA, we also evaluate the bilingually trained (Chinese and English) Yi models~\cite{ai2024yi} for the single-image VQA task.

\begin{table}[t]
    \centering
    \resizebox{\linewidth}{!}{
    \begin{tabular}{lcccc}
        \toprule
        Evaluation & \multicolumn{2}{c}{Multi-image VQA} & \multicolumn{2}{c}{Single-image VQA} \\
        \cmidrule(lr){2-3} \cmidrule(lr){4-5}
                   & ZH & EN & ZH & EN \\
        \midrule
        \rowcolor[gray]{0.95}
        Human     & 91.69   & 77.22$^\dagger$     & 74.41   & 46.53$^\dagger$     \\
        \midrule
        Phi-3-vision-4.2B & 29.03  &  33.75   & 42.58   & 44.53    \\
        Idefics2-8B  & \textbf{50.87}  & 41.69   & 46.87   & \textbf{52.73}     \\
        Mantis-8B    & 46.65  &  \textbf{43.67}   & 41.80   & 47.66    \\
        Qwen-VL-12B  & 32.26  &  27.54   & 48.83   & 42.97    \\
        Yi-VL-6B     & -      &  -   & \textbf{49.61}   & 41.41   \\
        \midrule
        Yi-VL-34B    & -      &  -  & 52.73   & 48.05     \\
        \midrule
        GPT-4V       & 78.92    & 69.23    & 63.67   & 60.16   \\  
        GPT-4o       & \textbf{86.35}     & \textbf{80.64}   & \textbf{72.66}   & \textbf{67.97}    \\
        \bottomrule
    \end{tabular}}
    \caption{Comparison of Multi-image and Single-image VQA Performance in Chinese and English. We report the best accuracy from four prompts. $^\dagger$: see footnote.\label{tab:VQA} \footnotemark}
\end{table}

The evaluation accuracy is reported in Table~\ref{tab:VQA}. Almost every open-weight model performs better on Single-image VQA than Multi-image VQA. We can observe that, for the bilingually trained models, i.e., Qwen-VL and Yi-VL, their performance is better when evaluated in Chinese. However, for the multilingual models, i.e. Phi-3, Idefics2, and Mantis-8B, their performance is better when evaluated in English. The best performing models are the API-based models from OpenAI. \footnotetext{Results with $^\dagger$ denote an estimate, calculated over 100 random samples, of human performance on the English Multi-Image and Single-Image VQA from one native speaker with no specialized knowledge of Chinese food culture.} 

\subsection{Models are Strong at Text QA}
We evaluate text question answering with a series of open-weight models, including Phi-3-medium-4k-instruct~\cite{abdin2024phi3}, Llama3-8B-Chinese~\cite{shenzhi_wang_2024}, Mistral-7B-Instruct-v0.3~\cite{shenzhi_wang_2024}, Yi-6B and 34B models~\cite{ai2024yi}, and Qwen2-7B-instruct~\cite{qwen2}, as well as API-based model GPT-4~\cite{achiam2023gpt}. 

Given that translatin is challenging and would likely introduce additional information and unfair comparison, we only evaluate the text questions in Chinese. For example, a famous Sichuan dish ``\begin{CJK}{UTF8}{gbsn}\xpinyin*{夫妻肺片}\end{CJK}'' can be translated to "couple's lung slices" if translate word by word, however it would be translated as "Sliced Beef and Ox Tongue in Chilli Sauce" by meaning. While the literal translation makes no sense, translation by meaning would hint the flavor and ingredients that are not included in its original Chinese name. 

\begin{figure}[t]
    \centering
    \includegraphics[width=\columnwidth, trim=0mm 8mm 0mm 7mm, clip]{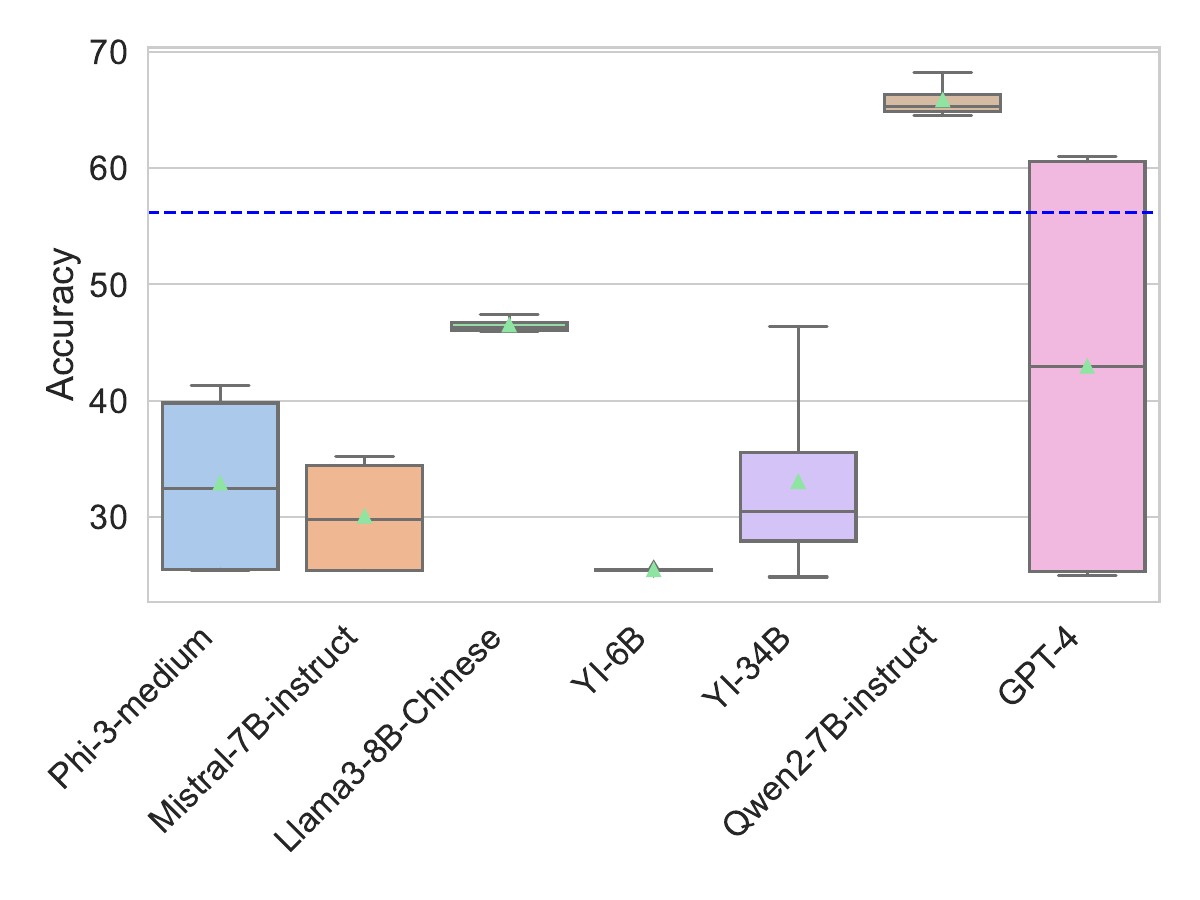}
    \caption{Accuracy of text QA across four different prompts. The blue dashed line indicates human accuracy~(56.2\%).}
    \label{fig:textqa-acc}
\end{figure}

From Figure~\ref{fig:textqa-acc}, we see that the Qwen2-7B-instruct model surpasses human performance on the text QA task, where the questions are formulated based on the local specialty annotations in Section~\ref{sec:dataset}. Since the local specialty annotations are collected and summarized by local representatives, potentially incorporating information from public web resources such as Baidu-Baike, the high performance may be attributed to the inclusion of domain-specific training data.

\begin{figure*}[t]
    \centering
    \begin{subfigure}[b]{0.48\textwidth}
        \centering
        \includegraphics[width=\textwidth, trim=0mm 5mm 0mm 5mm, clip]{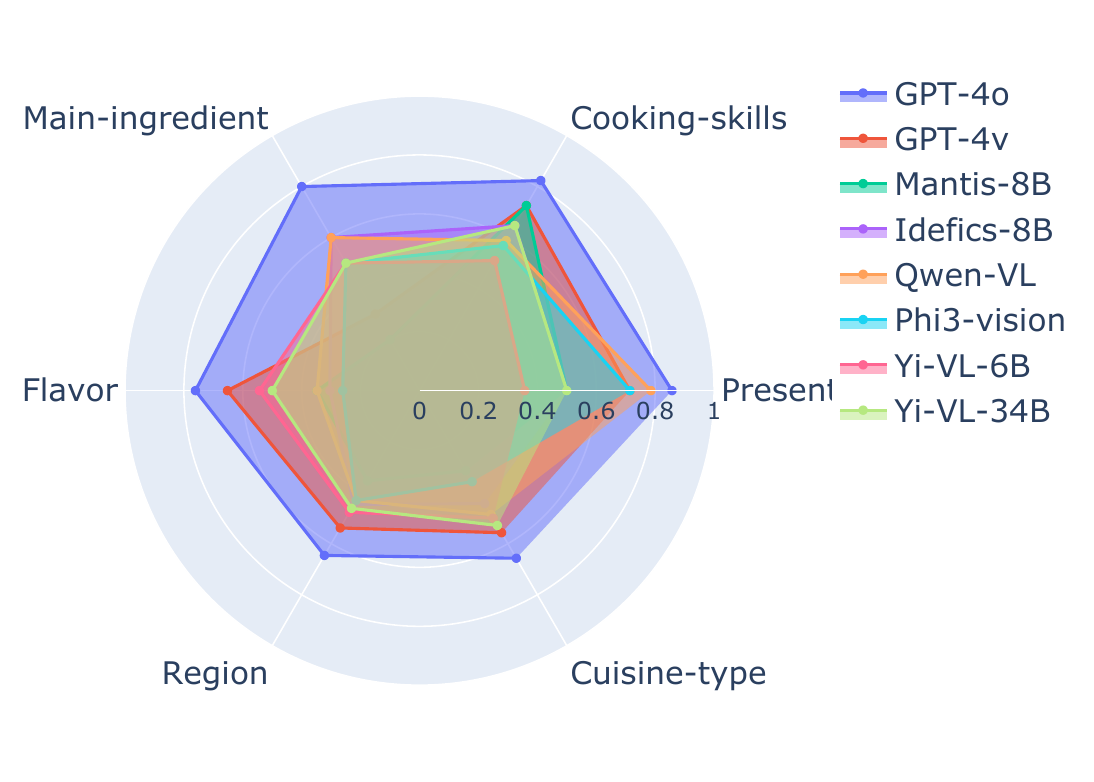}
        \caption{Single-image VQA.}
        \label{fig:radar-sivqa}
    \end{subfigure}
    \hfill
    \begin{subfigure}[b]{0.48\textwidth}
        \centering
        \includegraphics[width=\textwidth, trim=0mm 5mm 0mm 5mm, clip]{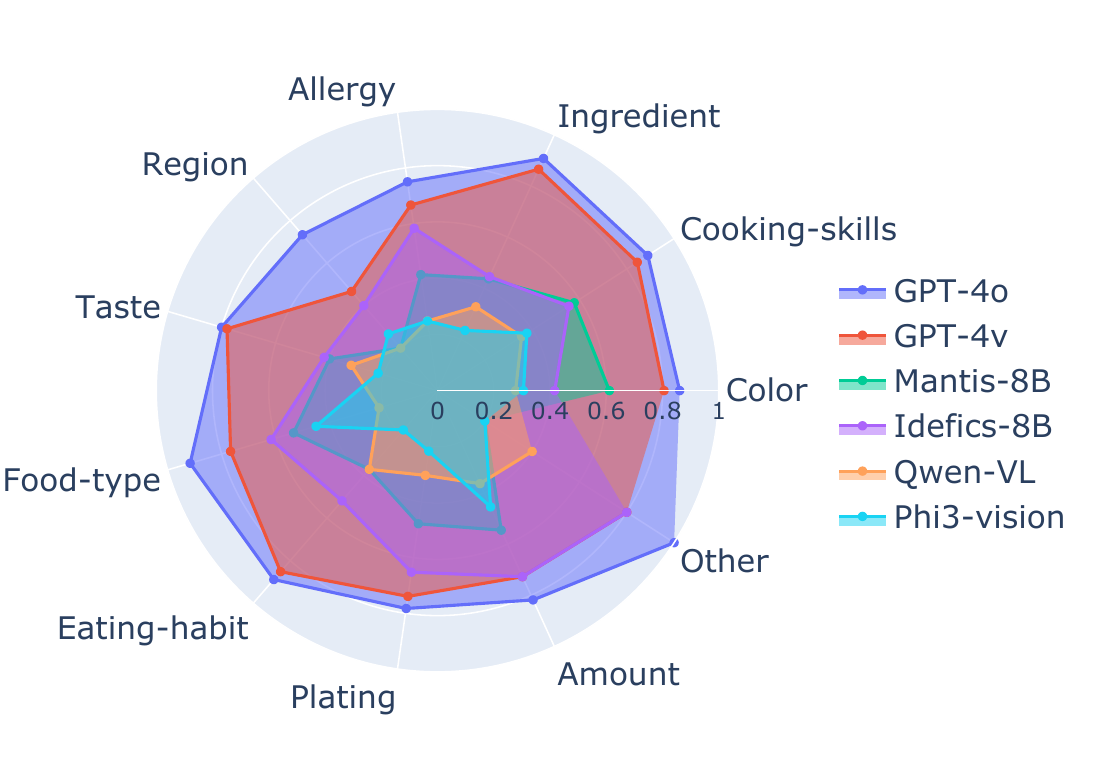}
        \caption{Multi-image VQA.}
        \label{fig:radar-mivqa}
    \end{subfigure}
    \caption{Model accuracy on fine-grained question attributes.}
    \label{fig:radar-vqa}
\end{figure*}

\section{Analysis}
\label{sec:analysis}

In this section, we explore which factors are important for fine-grained understanding of Chinese food culture.

\paragraph{Non-public images are crucial for fair evaluation.}

We incorporate user-uploaded non-public images into our dataset to prevent data contamination during evaluation. To verify the importance of preserving these non-public images for fair evaluation, we compare model performance using web-sourced images instead. Specifically, we manually searched with dish names to obtain web images for 171 out of 256 questions in the Single-image VQA task. As shown in Table~\ref{tab:ablate-web-img}, replacing non-public images with web-sourced dish images made the task easier for baseline models, indicating potential data contamination from web sources. Therefore, the use of non-public images is crucial for ensuring fair evaluation.

\begin{table}[h]
\centering
\resizebox{0.95\linewidth}{!}{
    \begin{tabular}{lcc}
    Model    & non-public images    & web images  \\
    \midrule
    Qwen-VL-12B   & 43.75 & \textbf{47.95}  \\
    Idefics2-8B   & 45.60 & \textbf{47.07}  \\
    Yi-VL-6B      & 47.56 & \textbf{50.88}  \\
    \bottomrule
    \end{tabular}}
\caption{
Models obtain higher accuracy when evaluating with web images, which indicates possible data contamination. The accuracy scores are averaged over four prompts. 
\label{tab:ablate-web-img}
}
\end{table}

\paragraph{Visual information helps.}

\begin{table}[t]
\centering
\resizebox{\linewidth}{!}{
    \begin{tabular}{lcccc}
      \toprule
      Input       & prompt1    & prompt2    & prompt3    & prompt4    \\
    \midrule
    Dish name only    & 28.52 & 27.73 & 36.72 & 37.11 \\
    + dish image   & 40.23 & 41.41 & 40.62 & 42.19 \\
    \bottomrule
    \end{tabular}}
\caption{
Accuracy on two variants of Single-image VQA task,
showing that visual information of food images is crucial for Idefics2 to correctly answer the questions.
\label{tab:ablate-img}
}
\end{table}

In Single-image VQA, the default setting is to query with only dish image without specifying the dish name.
We now examine whether the visual information is beneficial using the Idefics2-8B model.\footnote{We selected this model because it supports text-only inputs, unlike some other models such as the Yi-VL series.}
Results are shown in Table~\ref{tab:ablate-img},
where we investigate two variants:\
querying the model with only the text question but revealing the dish name, versus providing both the dish image and the dish name.
We observe that the Idefics2 model consistently performs better when dish images are available as visual clues. Please see comparison examples in Appendix~\ref{sec:app-compare}.

\paragraph{Dish names could be helpful clues for some of the models.}
\label{para:dishname}

As discussed in Section~\ref{para:qstats}, over 73.4\% of single-image questions require multi-hop reasoning, which typically involves identifying the dish and then leveraging related knowledge to answer the questions. To determine whether the identification of the food image and the utilization of visual information are bottlenecks for the models, we compare their performance on single-image VQA when provided with the dish name in the question.

The results in Table~\ref{tab:ablate-dishname} indicate that while the Yi models significantly benefit from being given both the images and names of the dishes, the Idefics2-8B model does not show the same improvement from this additional information. This indicates that recognizing the dishes could be a possible bottleneck for the Yi series models.

    \begin{table}[t]
        \centering
        \resizebox{\linewidth}{!}{
        \begin{tabular}{lcccccc}
              \toprule
              Model  & Condition & p1 & p2 & p3 & p4 \\
        \midrule
        \multirow{2}{*}{Yi-VL-6B}  & Image-only     & \textbf{49.61} & 48.05 & 47.66 & 46.09 \\
                                   & + dish name    & 73.83 & 74.61 & \textbf{76.17} & 62.50 \\
        \midrule
        \multirow{2}{*}{Yi-VL-34B} & Image-only     & 50.39 & \textbf{52.73} & 50.78 & 48.83 \\
                                   & + dish name    & 75.39 & 78.13 & \textbf{79.30} & 75.39 \\
        \midrule 
        \multirow{2}{*}{Idefics2-8B}& Image-only     & 44.53 & 43.75 & 46.09 & \textbf{46.87} \\
                                    & + dish name    & 40.23 & 41.41 & 40.62 & \textbf{42.19} \\
        \bottomrule
        \end{tabular}}
        \caption{Accuracy in the Single-image VQA task when dish name is revealed in the questions along with the image or not. While the Yi models benefit greatly from the additional information of the dish name, Idefics2 does not.
        ``p1--4'' indicates four different prompt templates.
        }
        \label{tab:ablate-dishname}
    \end{table}

\paragraph{Models are foodies who know cooking better than taste.}

Figure~\ref{fig:radar-sivqa} shows the model performance under fine-grained questions attributes on Single- and Multi-image VQA.
We observe that all models generally excel at answering questions related to cooking skills and ingredients. The Yi models, in particular, demonstrate a stronger ability to identify the flavors of dishes. Conversely, the Qwen-VL and Phi3-vision models perform well in observing the presentation of food when served but struggle with flavor-related questions. When answering questions based on multiple images, it also holds true that models are generally good at questions regarding cooking skills and the amount of food (Figure~\ref{fig:radar-mivqa}). However, these models are weak at answering questions related to the region and taste of the dish. Idefics-8B stands out, excelling in most of the fine-grained features we evaluated.

\begin{figure*}[t]
    \centering
    \begin{subfigure}[b]{0.42\textwidth}
        \centering
        \raisebox{4mm}{\includegraphics[width=\textwidth]{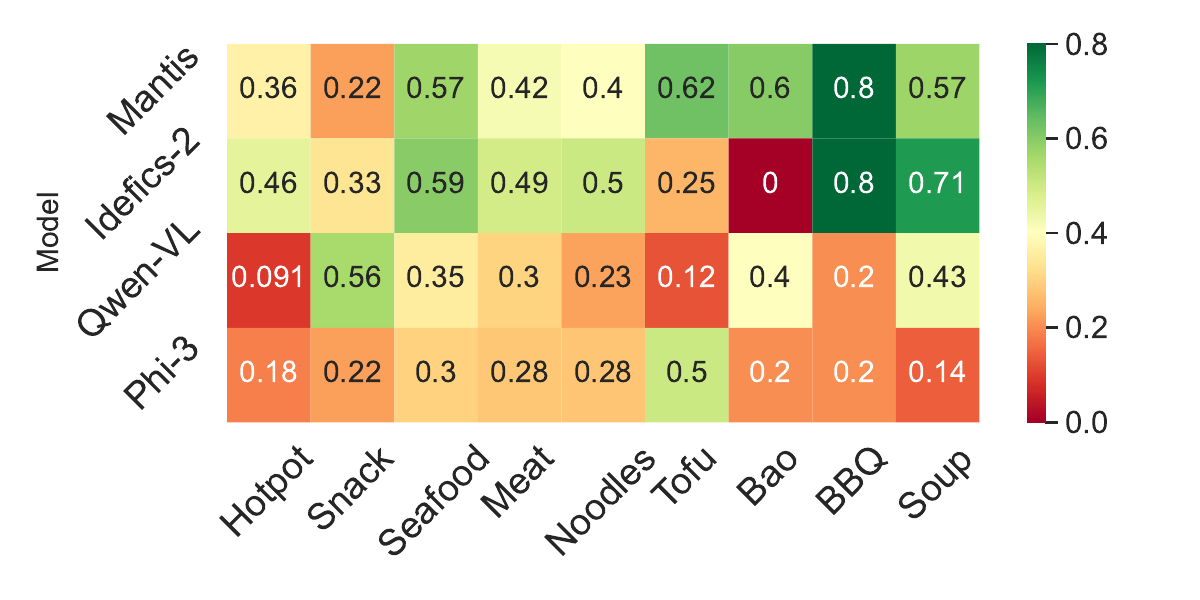}}
        \caption{Accuracy by food categories.}
        \label{fig:heatmap-food-type}
    \end{subfigure}
    \hfill
    \begin{subfigure}[b]{0.56\textwidth}
        \centering
        \includegraphics[width=\textwidth]{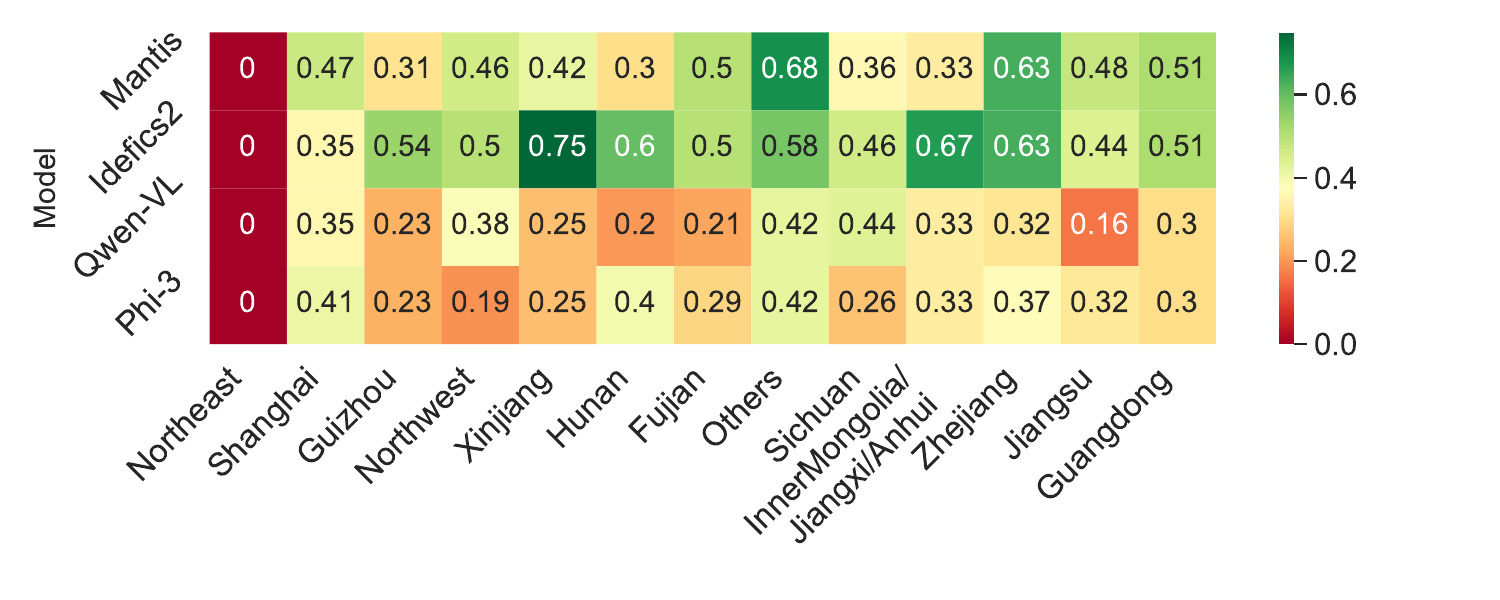}
        \caption{Accuracy by cuisine types.}
        \label{fig:heatmap-cuisine-type}
    \end{subfigure}
    \caption{Model accuracy on questions categorized by food categories and cuisine types.}
    \label{fig:heatmap-vqa}
\end{figure*}

\paragraph{Favorite food of the models.}
\label{para:heatmap}
In Figure~\ref{fig:heatmap-vqa}, we compare model performance on multi-image VQA tasks for questions grouped by food categories and cuisine types. This analysis provides insight into how well the models can compare features from images within the same group. 
The overall best performing model on multi-image VQA tasks excels at questions about BBQ and Xinjiang cuisines, but weak at questions about Shanghai dishes.
Another interesting finding is that, despite Sichuan food being one of the most popular cuisines in China, and presumably having more available images and resources online, none of the models excel at answering questions related to this cuisine type.

\section{Conclusion}
We introduce FoodieQA, a multimodal dataset designed to evaluate fine-grained understanding of Chinese food culture through multi-image, single-image, and text-only multiple-choice questions.

Our experiments, which focus on regional cultural differences and detailed visual features, reveal that understanding food and its cultural context remains a complex and under-explored task. We find that comparing food across multiple images—similar to the common scenario of people browsing menus—is particularly challenging. All open-source models underperform human accuracy by more than 40\% in this task. This suggests that our dataset offers a more accurate assessment of the suitability of state-of-the-art models for real-world applications in the food domain. 

Our analysis of language and prompt templates indicates that models can be sensitive to the language in which questions are asked---bilingually trained Chinese--English models perform better in Chinese, while other multilingual models are stronger in English. We also demonstrate the effectiveness of incorporating visual features compared to text-only settings in this context.

Improved models or methods for understanding food culture may be essential for future progress in the FoodieQA challenge. Looking ahead, we aim to expand the dataset to include dishes from other countries and regions. Following \citet{jacovi-etal-2023-stop}, we make our dataset a public benchmark on Huggingface at \href{https://huggingface.co/datasets/lyan62/FoodieQA}{lyan62/FoodieQA} with the \text{CC BY-NC-ND 4.0 License}. All of our data annotation and verification tools are freely available for re-use at \href{https://github.com/lyan62/FoodieQA}{github.com/lyan62/FoodieQA}. We encourage the community to create Foodie datasets for their own language and culture groups.

\section{Limitations}
The size of the FoodieQA dataset is limited by the challenge of collecting unseen images from individuals, as it requires them to voluntarily upload images from their phones or cameras. Although we have distributed the survey on two popular Chinese social media platforms, we anticipate that increased social media exposure or collaboration with food industry professionals could facilitate the collection of more images, and contribute to a training dataset for advancing this direction.

Translating Chinese dish names into other languages poses another challenge, as some dish names do not directly relate to their ingredients or cooking methods. Introducing translated dish names could potentially introduce additional information, leading to unfair comparisons among the models. Consequently, we have chosen to experiment solely with Chinese questions for the text-based queries.

We have benchmarked fifteen popular models using our dataset. However, due to the rapid advancements in the field, it is impossible to benchmark all trending models continuously. We hope our dataset will inspire future researchers to develop similar Foodie datasets for their own regions and languages, thereby guiding LLMs and VLMs towards a better understanding of regional food cultures.

\section*{Acknowledgements}
We are grateful to the volunteers for their generous contributions and efforts in providing high-quality food images that support our research. We extend our gratitude to Xi Liu, Yihe Zhang, Yu Sun, Yueyin Xu, Gefan Yang, Shixiong Wang, Penglong Ma, Daiwei Wang, Bo Cui, Yu Dong, Jinming Hu, Yufei Lin, and Zhongsheng Huang for serving as local experts. Their efforts in verifying and correcting the local specialty annotations and providing valuable feedback have been essential in ensuring the annotation's accuracy and completeness. We also thank Fengyuan Liu, Ruixiang Cui, Zhi Zhang, Yu Sun, and many of our friends and family who helped spread the image collection survey on social media for wide regional and group coverage. Special thanks to Jordan Boyd-Graber and Jimmy Lin for providing helpful research advice. Wenyan Li is supported by the Lundbeck Foundation (BrainDrugs grant: R279-2018-1145) and  a research grant (VIL53122) from VILLUM FONDEN. Jiaang Li is supported by Carlsberg Research Foundation (grant: CF221432). Li Zhou is supported by Shenzhen Science and Technology Research Fund (JCYJ20220818103001002) and Shenzhen Science and Technology Program (ZDSYS20230626091302006).

\bibliography{anthology,custom}
\bibliographystyle{acl_natbib}

\appendix

\section{Rule-based question formulation}
\label{sec:app-rules}
For text-based question answering we develop a rule-based question formulation method. For each question type, we have the meta information from the local specialty annotation (Section~\ref{subsec:dish-ann}). Then we design three to four templates for each of the question type. For example, for questions that ask about cuisine type, our templates include 
\begin{CJK}{UTF8}{gbsn}
\begin{itemize}
    \item <dish>是哪个地区的特色菜? (What region is <dish> a specialty dish of?)
    \item <dish>是哪个地区的特色美食? (In which region that <dish> is a local specialty?)
    \item 去哪个地方游玩时应该品尝当地的特色美食<dish>? Which place should you visit to taste the local specialty food <dish>?
\end{itemize}
\end{CJK}
Then, we randomly select cuisine types that are not the correct answer to serve as the alternative options. By utilizing different meta fields, we can generate multiple questions for each dish.

For single-image VQA, we associate the questions related to the dish with the corresponding dish image in our collection. We exclude questions of the warm-cold type—those that inquire whether a dish is served hot or cold—since these questions involve different dishes as options and are not suitable for the single-image scenario.

\section{Question type and answer distribution}
\label{sec:app-stats}

\begin{figure*}[h]
    \centering
    \includegraphics[width=0.7\linewidth, trim=1mm 1mm 1mm 1mm, clip]{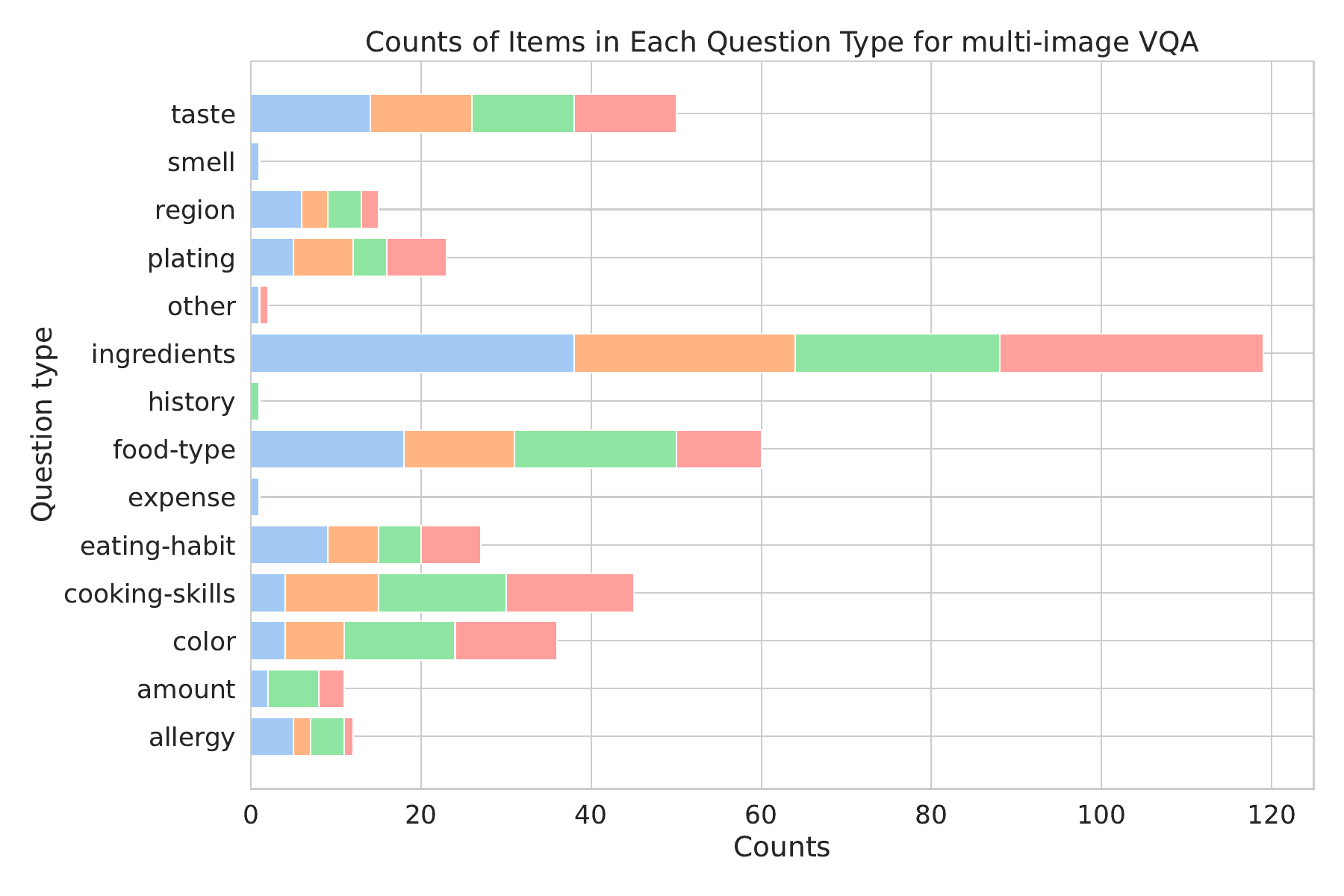}
    \includegraphics[width=0.7\linewidth, trim=1mm 1mm 1mm 1mm, clip]{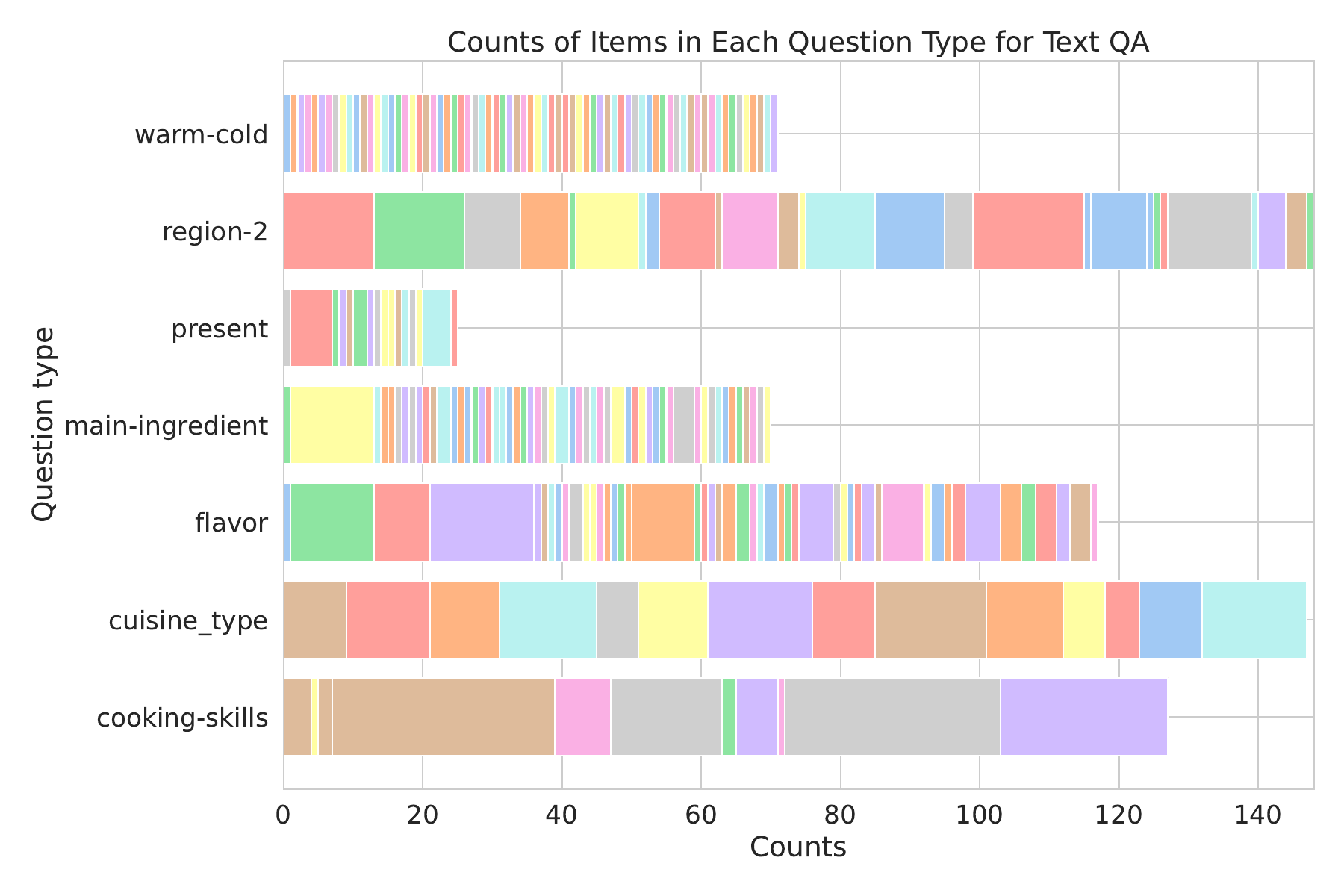}
    \includegraphics[width=0.7\linewidth, trim=1mm 1mm 1mm 1mm, clip]{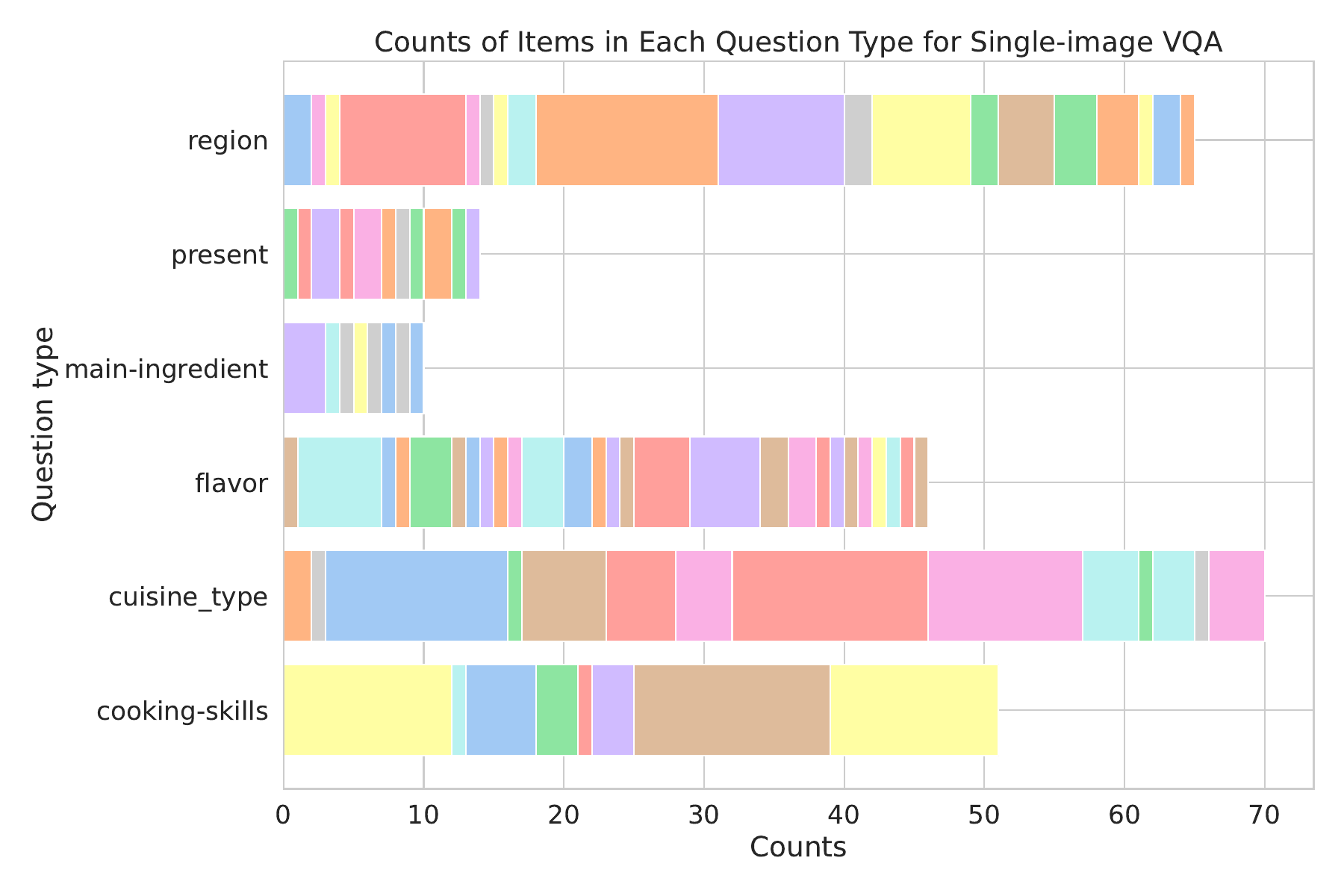}
    \caption{Answer distribution for each of the tasks. The questions are categorized by question type. Each color corresponds to a distinct answer, and each horizontal bar displays the distribution of these answers.\label{fig:ans-distr}}
\end{figure*}

In Table~\ref{tab:textqa-qdistr},~\ref{tab:sivqa-qdistr}, and~\ref{tab:mivqa-qdistr}, we show concrete statistics about distribution of question types in each task. Figure~\ref{fig:ans-distr} illustrates the answer distribution for questions categorized by type. Each horizontal bar independently displays the distribution of the answers regarding to the specific question type.

\begin{table}[h]
    \centering
    \begin{tabular}{lr}
        \textbf{Question type} & \textbf{Count} \\
        \midrule
        Cuisine Type       & 147 \\
        Cooking Skills     & 127 \\
        Main Ingredient    & 70  \\
        Region           & 148 \\
        Flavor             & 117 \\
        Present            & 25  \\
        Warm-Cold          & 71  \\
        \bottomrule
    \end{tabular} 
    \caption{Distribution of text QA question types.}
    \label{tab:textqa-qdistr}
\end{table}

\begin{table}[h]
    \centering
    \begin{tabular}{lr}
        \textbf{Question type} & \textbf{Count} \\
        \midrule
        Cuisine Type      & 70  \\
        Flavor            & 46  \\
        Region            & 65  \\
        Present           & 14  \\
        Cooking Skills    & 51  \\
        Main Ingredient   & 10  \\
        \bottomrule
    \end{tabular}
    \caption{Distribution of single-image VQA question types \label{tab:sivqa-qdistr}.}
\end{table}

\begin{table}[ht]
    \centering
    \begin{tabular}{lr}
        \textbf{Question type} & \textbf{Count} \\
        \midrule
        Ingredients        & 119 \\
        Food Type          & 60  \\
        Color              & 36  \\
        Taste              & 50  \\
        Cooking Skills     & 45  \\
        Plating            & 23  \\
        Eating Habit       & 27  \\
        Allergy            & 12  \\
        Region             & 15  \\
        Expense            & 1   \\
        Other              & 2   \\
        Amount             & 11  \\
        Smell              & 1   \\
        History            & 1   \\
        \bottomrule
    \end{tabular}
    \caption{Distribution of multi-image VQA question types \label{tab:mivqa-qdistr}.}
\end{table}

\section{Annotation Cost and Compensation}

\begin{table*}[h]
    \centering
    \begin{tabular}{lcc}
        \textbf{Task} & \textbf{Avg time/annotation} & \textbf{Avg time/person} \\ \hline
        Local specialty collection & 11.4 min/dish & 10.3 hrs/person \\
        Multi-image VQA question formulation & 3.5 min/question & 8.0 hrs/person \\
        Multi-image VQA question verification & 2.5 min/question & 6.7 hrs/person \\
        Single-image VQA verification & 3.3 min/question & 6.3 hrs/person \\ 
        TextQA verification & 1.2 min/question & 5.7 hrs/person \\ 
        \bottomrule
    \end{tabular}
    \caption{Average time per annotation and per person for annotation tasks. \label{tab:task-times}}
    
\end{table*}

In this work, the annotators are our colleagues who share co-authorship of the paper. This applies to the human annotation and validation process in Section~\ref{subsec:dish-ann}, Section~\ref{subsec:vqa}, and Section~\ref{subsec:human-validation}. The collection of images from private individuals, described in Section~\ref{subsec:image}, was entirely voluntary and by community effort through the social platforms, WeChat and Douban. 

The image collection period takes around one and a half months through the survey. Table~\ref{tab:task-times} displays an estimation of the annotation time reported by annotators.

\section{Prompts used for evaluation}
\label{sec:app-prompts}
Following \citet{durmus2023towards} and \citet{wang2024my}, we design four prompts for each of the tasks and extract the option letter from the model response. For multi-image VQA, we specifically include prompts that feature both interleaved image and text inputs as well as separate lists of images and texts. Please see examples of the prompts in Table~\ref{tab:prompts} and Table~\ref{tab:prompts2}.

\begin{CJK}{UTF8}{gbsn}

\newcommand{\pgeneral}{请从给定选项ABCD中选择一个最合适的答案。}

\begin{table*}[h!]
    \centering
    \begin{tabular}{@{} m{0.2\linewidth} m{0.75\linewidth} @{}}
    \toprule
    
    \textbf{Prompt 1} & 
    \begin{tabular}[t]{@{}m{0.8\linewidth}@{}}
    <img1>, <img2>, <img3>, <img4> \\
    根据以上四张图回答问题，他们分别为图A, 图B, 图C, 图D, \pgeneral 问题：<question>, 答案为：图 
    \end{tabular}\\
    \midrule
    \textbf{Prompt 2} & 
    \begin{tabular}[t]{@{}m{0.8\linewidth}@{}}
    <img1>, <img2>, <img3>, <img4> \\
    根据以上四张图回答问题, \pgeneral 问题：<question>, 答案为：图 
    \end{tabular}\\
    \midrule
    \textbf{Prompt 3} & 
    \begin{tabular}[t]{@{}m{0.8\linewidth}@{}}
    根据以下四张图回答问题, \pgeneral\\<img1>图A\\<img2>图B\\<img3>图C\\<img4>图D\\
    问题：<question>, 答案为：图 
    \end{tabular}\\
    \midrule
    \textbf{Prompt 4} & \begin{tabular}[t]{@{}m{0.8\linewidth}@{}}
    Human: 问题<question>，选项有: \\图A<img1>\\图B<img2>\\图C<img3>\\图D<img4>\\Assistant: 如果从给定选项ABCD中选择一个最合适的答案， 答案为：图 \end{tabular}\\
    \bottomrule
    \end{tabular}
    
    \caption{Chinese prompts for zero-shot evaluation for multi-image VQA. \label{tab:prompts}}

\end{table*}

\end{CJK}

\begin{table*}[h!]
\centering
\caption{Prompts for zero-shot evaluation}
\begin{tabular}{@{} m{0.2\linewidth} m{0.75\linewidth} @{}}
\toprule
\textbf{Prompt} & \textbf{Content} \\
\midrule
\textbf{Prompt 0} & 
\begin{tabular}[t]{@{}m{0.75\linewidth}@{}}
<img1><img2><img3><img4> \\
Answer the following question according to the provided four images, they correspond to Option (A), Option (B), Option (C), Option (D). Choose one best answer from the given options. \\
Question: {}, your answer is: Option (
\end{tabular} \\
\midrule
\textbf{Prompt 1} & 
\begin{tabular}[t]{@{}m{0.75\linewidth}@{}}
Answer the following question according to the provided four images which correspond to Option (A), Option (B), Option (C), Option (D). Choose one best answer from the given options. \\
The options are: \\
<img1>Option (A) \\
<img2>Option (B) \\
<img3>Option (C) \\
<img4>Option (D) \\
Question: <question>, your answer is: Option (
\end{tabular} \\
\midrule
\textbf{Prompt 2} & 
\begin{tabular}[t]{@{}m{0.75\linewidth}@{}}
Answer the following question according to the provided four images, and choose one best answer from the given options. \\
The options are: \\
<img1>Option (A) \\
<img2>Option (B) \\
<img3>Option (C) \\
<img4>Option (D) \\
Question: <question>, your answer is: Option (
\end{tabular} \\
\midrule
\textbf{Prompt 3} & 
\begin{tabular}[t]{@{}m{0.75\linewidth}@{}}
Human: Question <question> The options are: \\
Option (A)<img1> \\
Option (B)<img2> \\
Option (C)<img3> \\
Option (D)<img4> \\
Assistant: If I have to choose one best answer from the given options, the answer is: Option (
\end{tabular} \\
\bottomrule
\end{tabular}
\label{tab:prompts2}
\end{table*}

\section{Interface of image collection, annotation and verification tool}
\label{sec:ui}
\begin{figure*}[!t]
    \centering
        \includegraphics[width=0.9\textwidth, trim=1mm 1mm 1mm 1mm, clip]{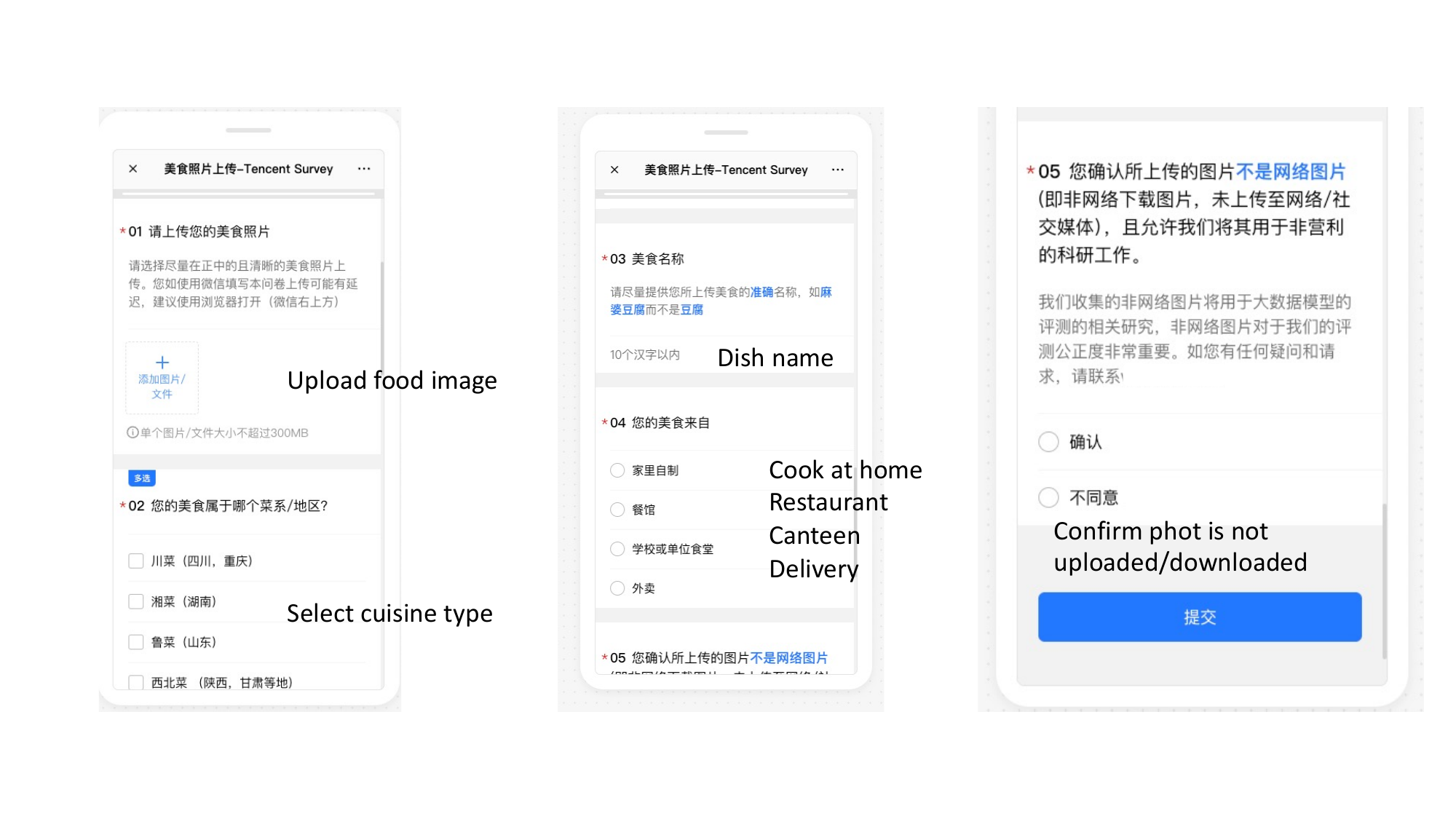}
        \caption{Survey interface of image collection}\label{fig:uploads}
    \end{figure*}

\begin{figure*}[t]
    \centering
        \includegraphics[width=0.9\textwidth, trim=1mm 1mm 1mm 1mm, clip]{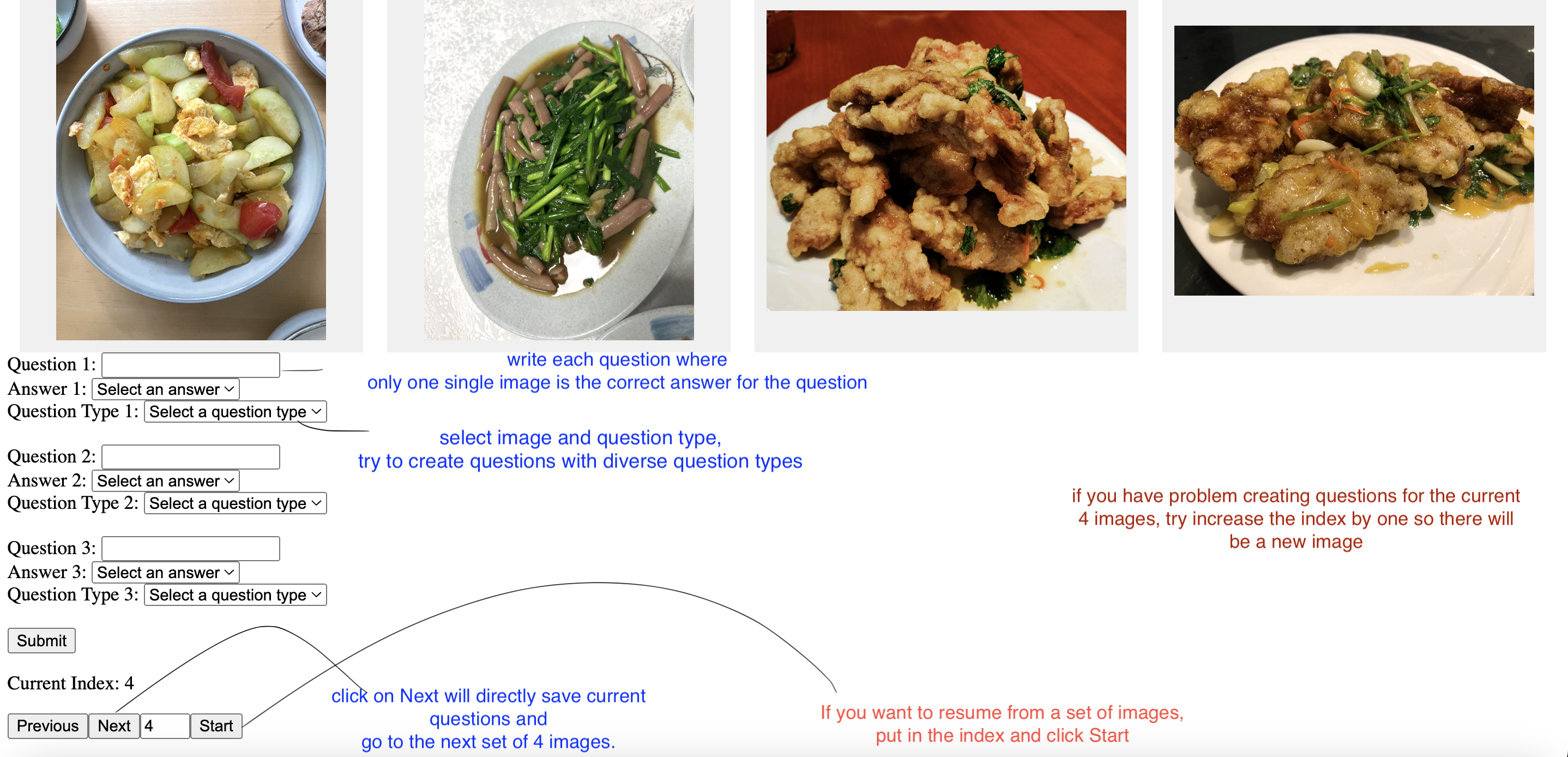}
        \caption{Annotation interface of writing questions when presented multiple images.}\label{fig:multi-vqa}
\end{figure*}

\begin{figure*}[t]
    \centering
        \includegraphics[width=0.9\textwidth, trim=1mm 1mm 1mm 1mm, clip]{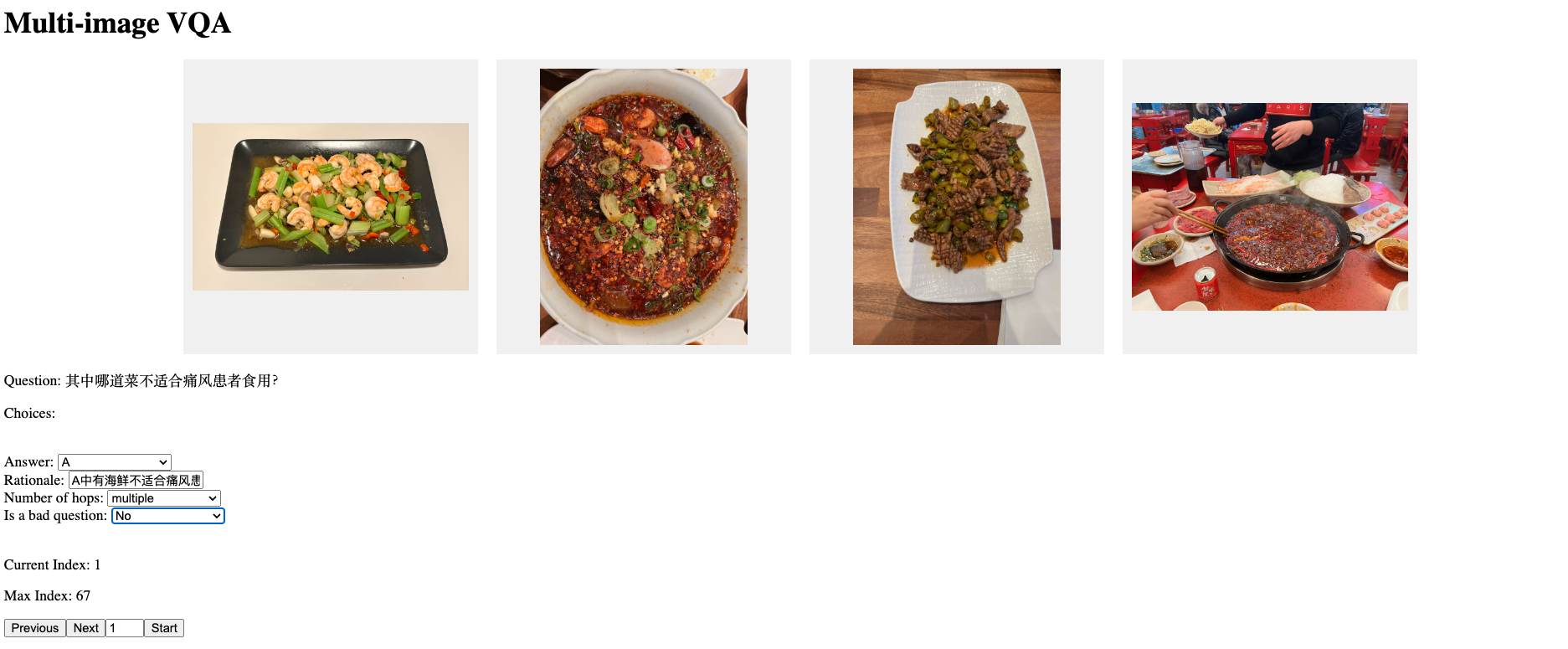}
        \caption{Annotation interface of verifying the multi-image multiple-choice questions.}\label{fig:multi-vqa-verify}
\end{figure*}

In Figure~\ref{fig:uploads}, we display the survey that we used to collect images. In Figure~\ref{fig:multi-vqa} and Figure~\ref{fig:multi-vqa-verify} show the user interface that annotators use to create questions and verify the questions.

\section{More examples}
\label{sec:app-ex}
\subsection{Examples of the questions in the dataset}
\begin{figure*}[t]
\centering
    \includegraphics[width=0.9\textwidth, clip]{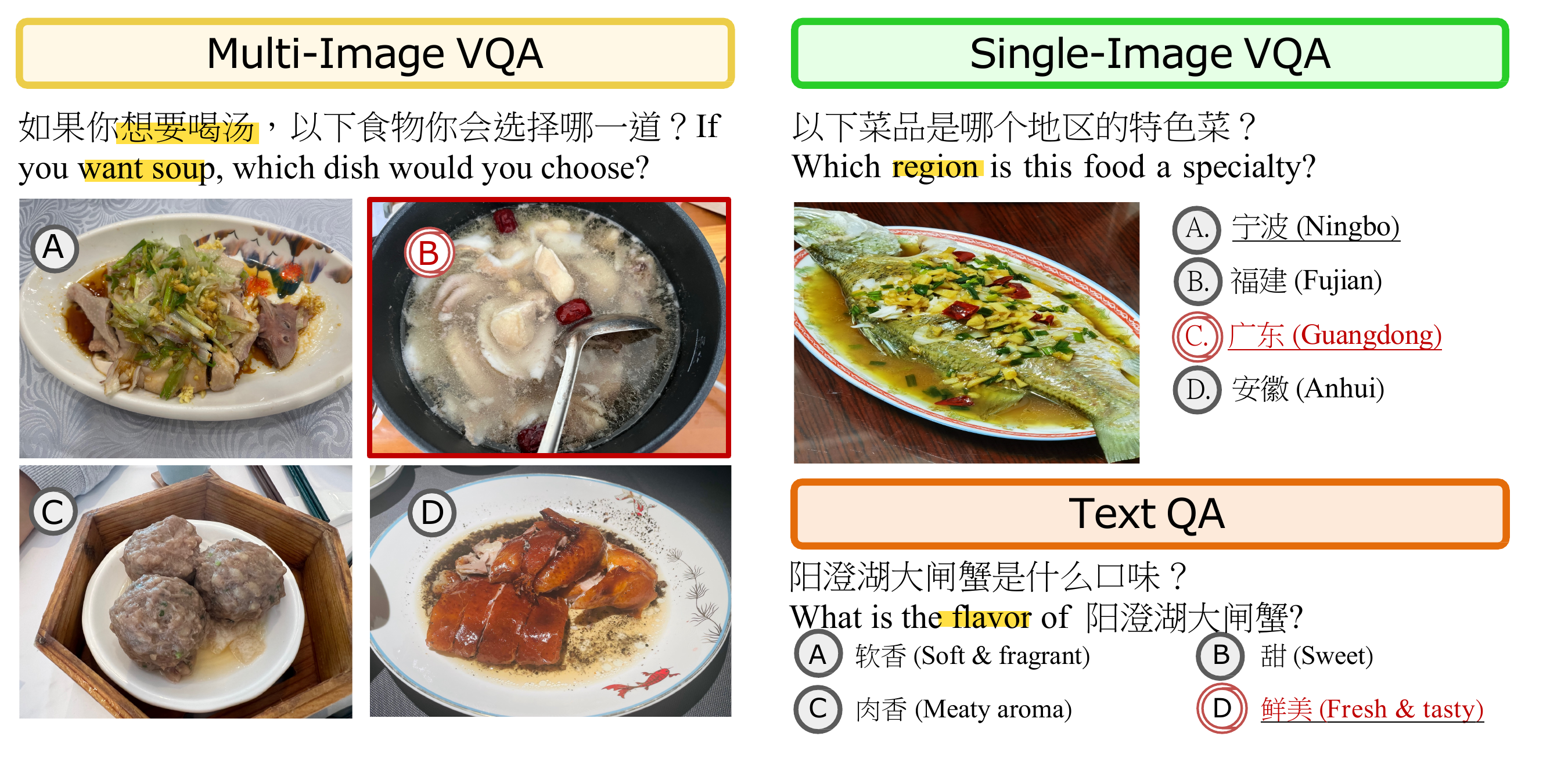}
    \includegraphics[width=0.9\textwidth, clip]{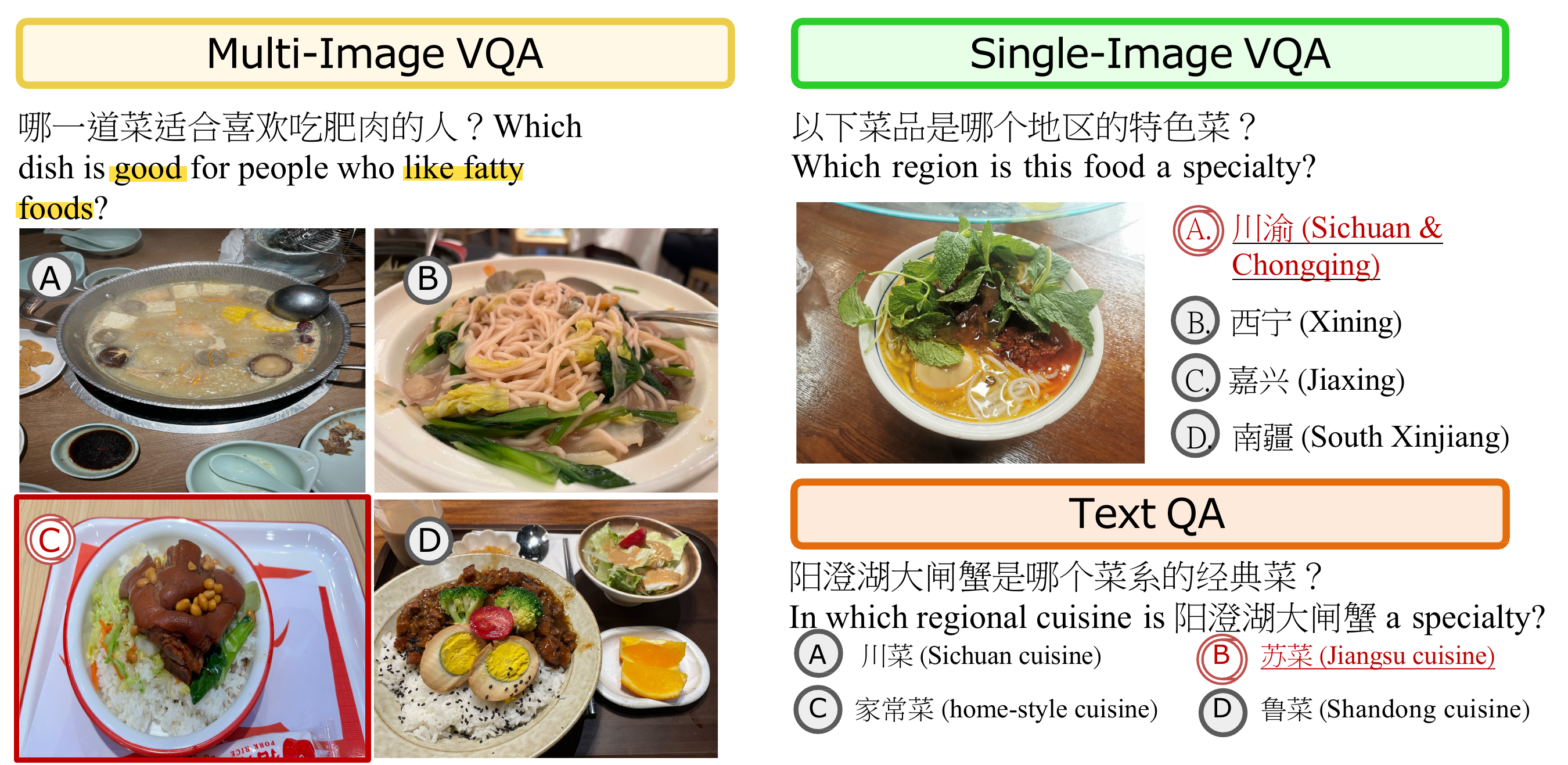}
    \includegraphics[width=0.9\textwidth, clip]{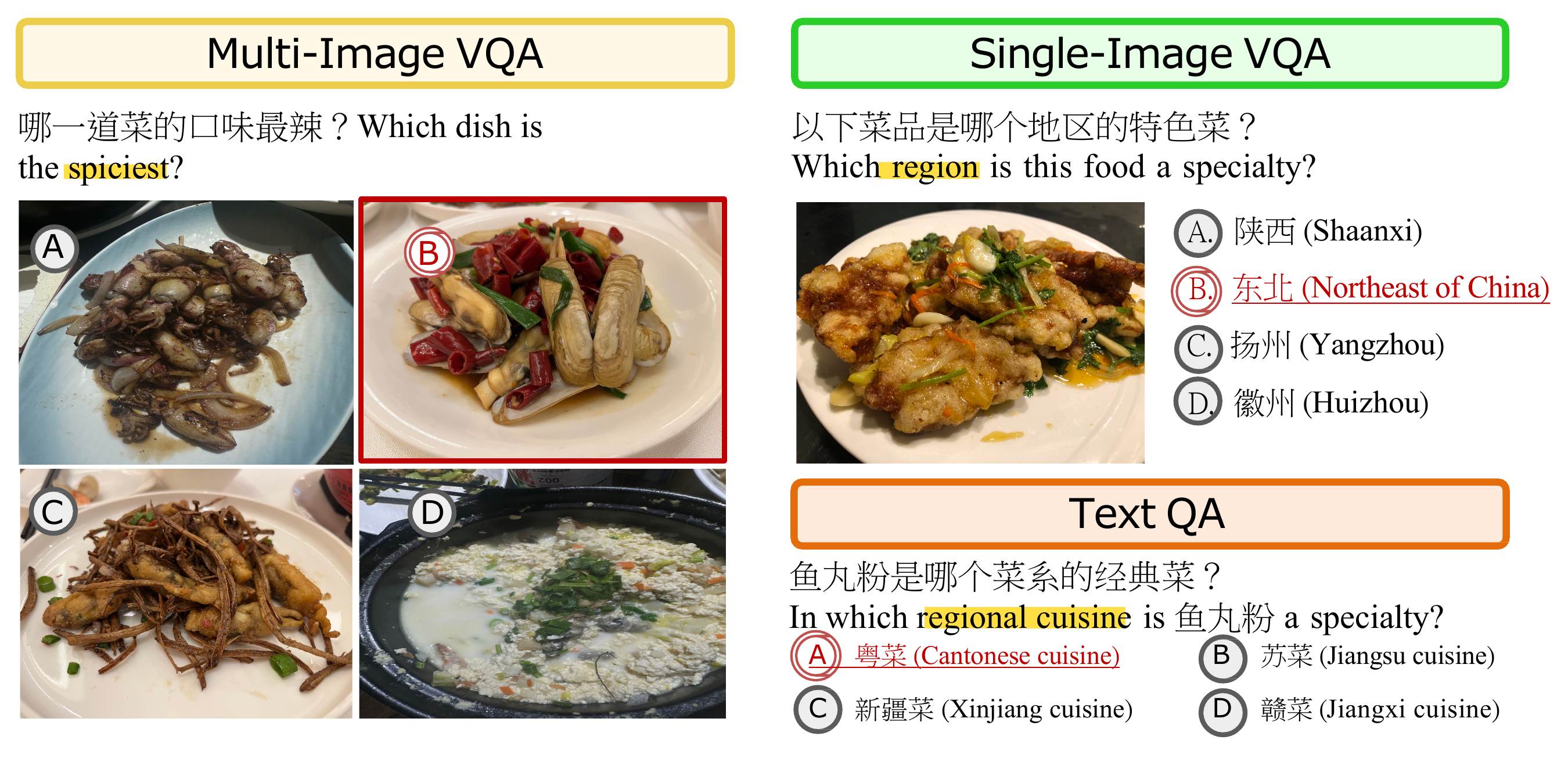}
    \caption{More examples in FoodieQA evaluate food culture understanding from three perspectives.}
\end{figure*}

\subsection{Examples of comparing whether the visual information is available}
\label{sec:app-compare}

\begin{figure}[t]
    \centering
        \includegraphics[width=\linewidth, trim=1mm 1mm 1mm 1mm, clip]{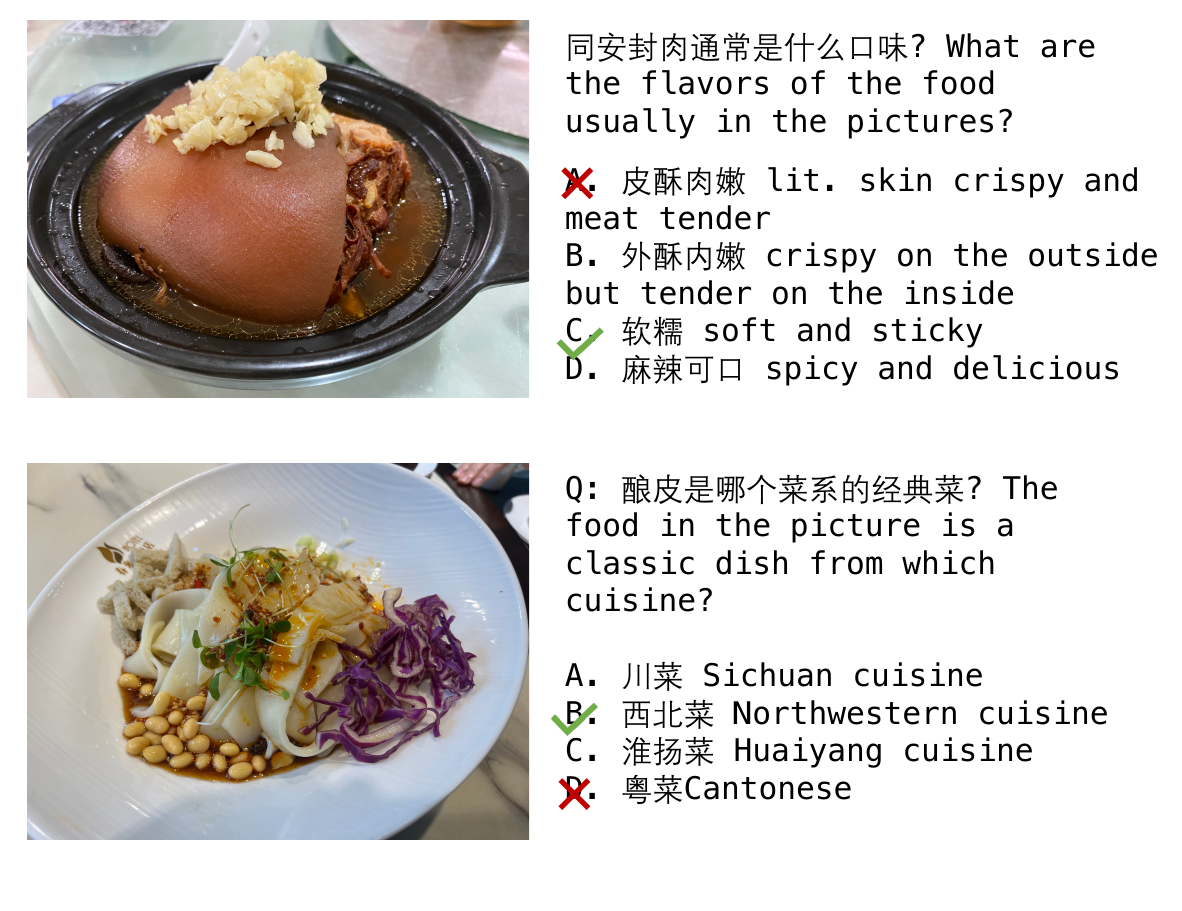}
        \caption{Examples where the Idefics-2-8B model correctly answers the question when the image is available but failed when it is not.}\label{fig:show-imag}
    \end{figure}

\end{document}